\documentclass[acmlarge,screen,nonacm]{acmart}
\makeatletter

\let\@ACM@checkaffil\relax
\makeatother
\usepackage{amssymb}
\usepackage{placeins}
\usepackage{graphicx}%
\usepackage{bbding}%
\usepackage{pifont}%
\usepackage{multicol}
\usepackage{nicematrix} %
\usepackage{tikz}

\usepackage{fontawesome}

\usepackage{stmaryrd} %

 \usepackage[flushleft,para]{threeparttable}
\usepackage{pifont}%
\newcommand{\cmark}{\ding{51}}%
\newcommand{\xmark}{\ding{55}}%
\usepackage{rotating}
\newcommand*{\headformat}[1]{#1}
\newlength{\maxlen}
\newcommand*{\shortheadformat}[1]{#1}
\newlength{\shortmaxlen}
\AtBeginDocument{%
  }

\setcopyright{none}
\acmDOI{}
\acmISBN{}

\usepackage{booktabs} %
\usepackage{graphicx} %
\usepackage{lipsum}   %
\usepackage{amsmath,amssymb,amsfonts}
\usepackage[noend]{algorithmic}
\usepackage{algorithm}
  
\usepackage{float}
\usepackage{subfig}
\usepackage{multirow} 
\usepackage{textcomp} 
\usepackage{setspace}
\usepackage{bm}

\usepackage{makecell}
\begin{document}

\title[VAFL: Vertical Analytic Federated Learning]{VAFL: A Vertical Analytic Federated Learning Mechanism for Efficient Crowdsourced Services via Closed-Form Solutions}

\author{Yajiang Huang}
\affiliation{%
  \institution{Central South University}
  \country{China}
}

\author{Anfeng Liu}
\affiliation{%
  \institution{Central South University}
  \country{China}
}

\author{Tian Wang}
\affiliation{%
  \institution{Beijing Normal University}
  \country{China}
}

\author{Mianxiong Dong}
\affiliation{%
  \institution{Muroran Institute of Technology}
  \country{Japan}
}

\author{Houbing Herbert Song}
\affiliation{%
  \institution{University of Maryland, Baltimore County}
  \country{USA}
}

\renewcommand{\shortauthors}{Huang et al.}

\begin{abstract}
Intelligent services often rely on complementary information held by different organizations, motivating Vertical Federated Learning without centralized raw-data pooling. We propose \textbf{VAFL}, a Vertical Analytic Federated Learning framework that fits classification and regression heads on fixed local representations through regularized least squares. An agent-assisted protected-computation workflow supports two-party and multi-party collaboration. To reduce fitting overhead, we develop a full-dimensional equivalent solve that applies block factors directly to the target matrix, avoiding construction of the full inverse. We further study two width-reduction strategies: local spectral projection with coefficient lifting, and full-dimensional local fitting with low-rank cross-party correction. The first compresses the prediction space, whereas the second retains local components outside the collaboration subspace. We establish their algebraic properties and analyze their computational scope. Experiments on five benchmarks show the accuracy--cost tradeoffs of these strategies and compatibility with diverse frozen backbones. Predictive performance and training time identify useful operating points and the limits of compression.
\end{abstract}


\maketitle

\section{Introduction}

With the rapid development of mobile and ubiquitous computing, intelligent services have become an integral part of everyday life. Personalized content recommendation, mobile commerce, financial risk assessment, and healthcare applications increasingly rely on data generated through users' interactions with digital platforms and connected devices~\cite{app-1,app-2,app-3,app-4,app-5,app-6}. Collectively, these crowd-generated data support more personalized and responsive services. In this sense, such services are \emph{crowd-powered}: their data originate from users' daily activities, their intelligence is built upon aggregated contributions, and their benefits return to individual users. This relationship---\emph{data from the crowd, intelligence through collaboration, and services for the crowd}---motivates efficient collaborative learning for user-oriented ubiquitous services \cite{tang2026eqtptd, fan2026pcqsp, tang2026qdbpp, acu, huang2025qlpdcs, tang2025cpdz}.

Building these services requires complementary information held by different providers. For example, a short-video platform may observe viewing histories, while an e-commerce provider and a mobile operator hold purchase preferences and service-usage information. Similarly, financial prediction services may benefit from combining transaction features with other permitted user attributes maintained by separate institutions. Yet such data are fragmented across application providers, mobile operators, banks, and other organizations. Integrating these complementary perspectives while keeping individual records with their respective holders raises organizational and technical challenges.

\begin{figure}[t]
    \centering
    \includegraphics[width=1.0\linewidth]{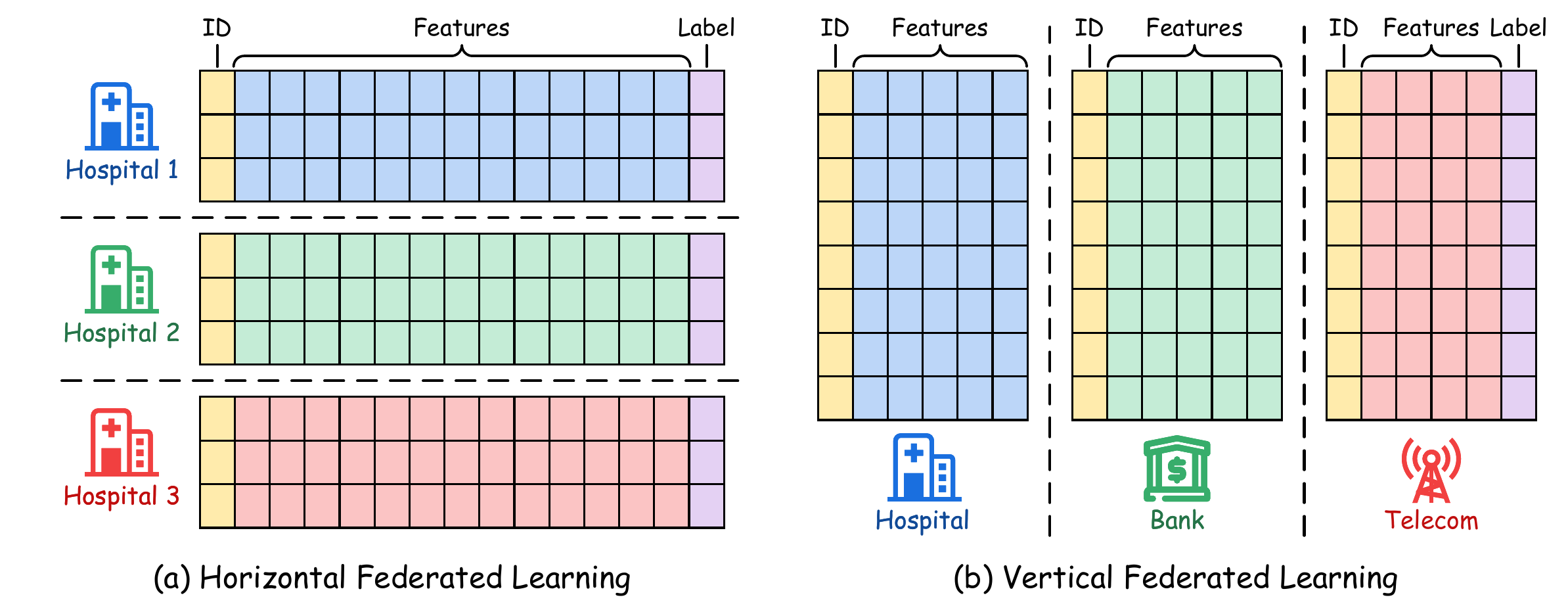}
    \caption{Comparison of HFL and VFL.}
    \label{fig:introduction-vfl}
    \vspace{-0.45cm}
\end{figure}

Privacy and data ownership concerns further complicate collaboration. User behavior, application usage, and financial activities may reveal sensitive personal information, while regulations such as the General Data Protection Regulation (GDPR) constrain data processing and sharing~\cite{privacy-1,privacy-2,privacy-3,TSC-2}. Federated Learning (FL) enables collaborative training without directly pooling raw records~\cite{FL-survery,HFL-survey-2,CALM}. Based on data partitioning, FL includes Horizontal Federated Learning (HFL)~\cite{HFL-1,HFL-2,HFL-survey-1}, Vertical Federated Learning (VFL)~\cite{VFL-survey-0,VFL-survey-1,VFL-survey-2}, and Federated Transfer Learning (FTL)~\cite{FTL-1,FTL-2,FL-survery}. As illustrated in Fig.~\ref{fig:introduction-vfl}, HFL involves different samples sharing a feature space, whereas VFL connects complementary features held by organizations for aligned individuals~\cite{VFL-survey-0,Linear-1,VFL-survey-2}. VFL is therefore well suited to cross-platform services in which users generate data but organizations serve as collaborating parties, including healthcare, finance, recommender systems, and the Industrial Internet of Things (IIoT)~\cite{app-5,app-6,app-1,app-2,app-3,app-4,app-7,app-8,app-9}.

Despite its potential, VFL can incur substantial training overhead as aligned user records and complementary feature representations grow. Tree ensembles and split neural networks are widely used in VFL~\cite{Boost-1,Boost-2,Boost-3,Boost-4,NN-1,NN-2,NN-3}. Tree-based methods repeatedly evaluate protected split statistics, while split neural networks exchange activations and gradients across minibatches and epochs. When encryption or other protected computation is involved, these interactions introduce considerable computation and communication costs. Reducing collaborative fitting overhead is therefore important for constructing and updating data-intensive intelligent services.

Analytic Learning (AL) offers an alternative by fitting prediction layers through least-squares or related matrix solutions without iterative gradient optimization~\cite{GACL,CALM,new-AL-1, CFSSeg}. It has been studied in HFL~\cite{Fed3R,AFL,AFCL,AnaPFL}, continual learning~\cite{new-AL-2,new-AL-3,new-AL-4}, and crowdsourced service construction~\cite{CALM}. Unlike HFL, where local feature products can be combined in a shared feature space, VFL requires off-diagonal products across distinct feature blocks. Secure linear and ridge-regression methods already support direct solving, including recursive fitting across multiple feature holders~\cite{rw-gascon-2017,Linear-0}. These results provide an important foundation, but wide local representations make protected cross-party products and matrix solving expensive. The question we address is therefore: \emph{how can multiple data holders fit analytic prediction heads on complementary representations while reducing protected computation and retaining useful predictive information?} Section~\ref{Preliminaries} analyzes these cross-feature computations.

To address this challenge, we propose \textbf{VAFL}, a \textbf{V}ertical \textbf{A}nalytic \textbf{F}ederated \textbf{L}earning framework for collaborative prediction. VAFL uses vertically recursive regularized least squares to incorporate cross-party feature products and recover the centralized analytic solution for the same representations, targets, and regularization in exact arithmetic. It supports one-hot classification and scalar regression targets, with inputs comprising numerical descriptors or frozen CNN and transformer representations. For protected matrix computation in two-party and multi-party settings, VAFL combines Paillier encryption, secret sharing, and local perturbations in an agent-assisted workflow. Its privacy properties and limitations are discussed in Section~\ref{subsec:Privacy Analyses}.

To improve fitting efficiency, we introduce complementary strategies for exact computation and reduced collaboration. An equivalent full-dimensional solution procedure avoids constructing a complete inverse when only the prediction coefficients are required. For further savings, local dimensionality reduction decreases the width of protected computation. We also develop a strategy that retains full-dimensional local fitting while reducing cross-party interactions to a smaller subspace. Together, these designs allow the amount of collaborative computation to be adjusted to the available resources and representations. The equivalent procedure preserves the original analytic solution in exact arithmetic. Reduced collaboration introduces an approximation whose effect on prediction must be assessed.

For crowd-powered services, VAFL offers complementary efficiency benefits. Its non-iterative head fitting avoids repeated gradient updates and learning-rate or epoch selection. At fixed feature width, additional records change the feature products without enlarging the feature-space system to be solved, while reduced collaboration addresses the cost of wide representations. Computing feature products still depends on sample count, and retaining local information does not guarantee unchanged predictive accuracy. These properties and limits motivate our study of accuracy and fitting cost for privacy-conscious, cross-platform services.

The main contributions are as follows:
\begin{enumerate}
    \item \textbf{Analytic VFL framework.} We develop a collaborative prediction-head framework for classification and regression over complementary feature blocks. It accommodates numerical descriptors and frozen neural representations, with two-party and multi-party execution. The framework builds on analytic least squares and established feature-block recursion to support privacy-conscious, crowd-powered services.

    \item \textbf{Strategies for efficient analytic collaboration.} We introduce an equivalent solution procedure and complementary strategies that reduce the dimensions of protected computation. These designs distinguish computation that can be reorganized without changing the solution from collaboration that can be compressed, and retain local predictive components outside the reduced collaboration space.

    \item \textbf{Theoretical and privacy analyses.} We establish full-dimensional equivalence to centralized regularized least squares and characterize the solutions and approximation errors of the reduced strategies. We analyze computation and communication costs and message-level privacy under our assumptions.

    \item \textbf{Comprehensive experimental evaluation.} We compare predictive performance against representative baselines on classification and regression benchmarks, and examine fitting cost through dedicated scaling and acceleration experiments. The evaluation also covers regularization sensitivity and frozen-backbone compatibility. These experiments identify useful accuracy--cost tradeoffs and the limits of aggressive compression across different representations.
\end{enumerate}

The remainder of this paper is organized as follows. Section 2 reviews related work, Section 3 analyzes analytic learning under horizontal and vertical partitioning, Section 4 presents VAFL, Section 5 reports the experiments, and Section 6 concludes the paper.

\begingroup
\setlength{\textfloatsep}{12pt}
\setlength{\intextsep}{12pt}
\section{Related Work}
\label{Related Work}

We review representative VFL methods and analytic-learning approaches. Table~\ref{tab:related-work-comparison} compares their problem settings, fitting properties and approaches to reducing collaborative computation. The number of parties counts feature holders in vertical methods and sample holders in horizontal methods, excluding auxiliary computing agents. Cross-party products refer to Gram blocks computed between feature holders for analytic fitting. Feature-block recursion denotes an extensible recurrence across feature holders, beyond a fixed two-block solve. Non-iterative fitting denotes direct analytic head estimation without iterative parameter optimization, given fixed input representations. Frozen-backbone evaluation and modalities describe the studies reported in the compared work. We distinguish a projected head, whose coefficients are confined to reduced local subspaces and can be lifted to the supplied coordinates, from retaining full-dimensional local fits while reducing cross-party coupling. Check marks denote documented properties, while crosses denote properties outside the compared formulation or evaluation and do not rule out possible extensions. The acceleration properties summarize the options available within a method, rather than requiring them in every configuration.

\begin{table}[!htbp]
\centering
\caption{Comparison with related methods: problem settings, fitting properties and reduction of collaborative computation. C \& R denotes coverage of both classification and continuous-target regression in the compared work.}
\label{tab:related-work-comparison}
\begingroup\normalfont\small
\setlength{\heavyrulewidth}{0.6pt}
\setlength{\lightrulewidth}{0.4pt}
\setlength{\tabcolsep}{1.1pt}
\renewcommand{\arraystretch}{1.22}
\begin{tabular*}{\linewidth}{@{\extracolsep{\fill}}llclccccccc@{}}
\toprule
\textbf{Method} & \textbf{Partition} & \textbf{Parties} & \textbf{Modalities} & \textbf{C \& R} & \makecell{\textbf{Non-}\\\textbf{iterative}\\\textbf{fitting}} & \makecell{\textbf{Cross-}\\\textbf{party}\\\textbf{products}} & \makecell{\textbf{Feature-}\\\textbf{block}\\\textbf{recursion}} & \makecell{\textbf{Frozen}\\\textbf{backbone}\\\textbf{evaluation}} & \makecell{\textbf{Projected}\\\textbf{head}} & \makecell{\textbf{Local fit +}\\\textbf{reduced}\\\textbf{coupling}} \\
\midrule
Fed3R~\cite{Fed3R} & Samples & -- & Images & \xmark & \cmark & \xmark & \xmark & \cmark & \xmark & \xmark \\
AFL~\cite{AFL} & Samples & -- & Images & \xmark & \cmark & \xmark & \xmark & \cmark & \xmark & \xmark \\
AFCL~\cite{AFCL} & Samples & -- & Images & \xmark & \cmark & \xmark & \xmark & \cmark & \xmark & \xmark \\
CALM~\cite{CALM} & Samples & -- & Images & \xmark & \cmark & \xmark & \xmark & \cmark & \xmark & \xmark \\
\midrule
PyVertical~\cite{NN-1} & Features & $2$ & Images & \xmark & \xmark & \xmark & \xmark & \xmark & \xmark & \xmark \\
SecureBoost~\cite{Boost-2} & Features & $\ge2$ & Tabular & \cmark & \xmark & \xmark & \xmark & \xmark & \xmark & \xmark \\
Secure LR~\cite{rw-gascon-2017} & Features & $\ge2$ & Tabular & \xmark & \cmark & \cmark & \xmark & \xmark & \xmark & \xmark \\
VRL-2P~\cite{Linear-0} & Features & $2$ & Tabular & \xmark & \cmark & \cmark & \xmark & \xmark & \xmark & \xmark \\
VRL-MP~\cite{Linear-0} & Features & $\ge2$ & Tabular & \xmark & \cmark & \cmark & \cmark & \xmark & \xmark & \xmark \\
\midrule
\textbf{VAFL (Ours)} & Features & $\ge2$ & \makecell[l]{Tabular,\\images} & \cmark & \cmark & \cmark & \cmark & \cmark & \cmark & \cmark \\
\bottomrule
\end{tabular*}
\endgroup
\end{table}

\subsection{Vertical Federated Learning}

VFL allows institutions to collaborate on aligned records while holding different feature subsets locally~\cite{VFL-survey-0,VFL-survey-1,VFL-survey-2}. Its model families include tree ensembles, split neural networks and linear predictors, with different computation and communication requirements. SecureBoost~\cite{Boost-2} constructs boosting trees across multiple feature holders using encrypted gradient and Hessian statistics. Its formulation covers classification and regression, with successive tree-building stages. PyVertical~\cite{NN-1} uses a multi-headed split neural network and exchanges intermediate activations and gradients for back-propagation. Its demonstration uses two feature owners and a label-holding data scientist, and discusses extension to additional owners. These methods provide representative alternatives to solving a fixed-feature prediction head directly.

Linear and ridge-regression methods are particularly close to VAFL because they also exploit feature and target products. Existing work includes coreset-based regularized regression~\cite{Linear-1} and privacy-oriented linear or logistic models~\cite{Linear-2,Linear-3,Linear-4,Linear-5,Linear-6}. Direct vertical solvers also exist. Gasc{\'o}n et al.~\cite{rw-gascon-2017} study secure linear regression with multiple data providers and evaluate Cholesky-based direct solving alongside conjugate-gradient alternatives. Their protocol forms additive shares of the normal-equation terms and uses a garbled-circuit solving phase. The Secure LR entry uses their direct Cholesky variant for the non-iterative fitting comparison. Cai et al.~\cite{Linear-0} develop VRL-2P and VRL-MP for two-party and multi-party ridge regression. Their released implementation uses direct matrix operations, with VRL-MP recursively expanding the inverse across feature holders and exchanging encrypted and masked matrix quantities. Table~\ref{tab:related-work-comparison} separates the two variants to make their participant scope explicit. Accordingly, gradient-free solving, support for more than two holders and feature-block recursion are shared with relevant regression approaches.

RidgeLog-V2~\cite{RidgeLog-V2} addresses ridge logistic regression on vertically partitioned data. The work reports one exchange in each direction between the systems holding predictors and the response holder. It obtains the same parameter estimates as centralized fitting, with private responses and penalty cross-validation requiring no additional communication. This makes it a relevant comparison for communication-efficient vertical prediction. The reported communication efficiency does not by itself establish non-iterative parameter fitting.

VAFL studies analytic fitting as a common prediction-head framework for classification and regression on supplied local representations. Its scope includes numerical descriptors and vertically partitioned image features from frozen CNN and transformer backbones. It further introduces strategies for reducing protected computation, including collaboration in a reduced space while retaining full-dimensional local fitting. Their accuracy--cost tradeoffs are evaluated across feature budgets and participant counts. This extends the scope of the compared regression formulations without treating direct solving or multi-party recursion as new properties. The comparison concerns model and solution properties. Cryptographic guarantees depend on each method's observations and threat assumptions. VAFL's retained guarantee scope is stated in Section~\ref{subsec:Privacy Analyses}.

\subsection{Analytic Learning}

Analytic learning fits prediction layers through least-squares or related matrix solutions, removing gradient-based optimization of those layers~\cite{AL_1,AL_2,AL_3,ACIL,GACL}. Nonlinear representations can be supplied by a fixed feature extractor or mapping, while recursive identities allow sufficient statistics and solutions to be updated as data arrive. This provides a common mathematical basis for federated and continual-learning variants~\cite{Fed3R,AFL,AFCL,AnaPFL,CALM}. The head still depends on representation and regularization choices examined in our experiments.

Fed3R~\cite{Fed3R} fits a ridge classifier on frozen pretrained features in sample-partitioned FL. Clients compute local feature and target statistics, and the server combines them to recover the corresponding centralized ridge solution independently of client order and statistical heterogeneity. Its optional Fed3R-RF variant expands the representation through random features to approximate a kernel, rather than compressing feature blocks for vertical collaboration. Table~\ref{tab:related-work-comparison} compares the non-iterative Fed3R stage, excluding the subsequent gradient-based optimization in Fed3R+FT.

AFL~\cite{AFL} performs single-round federated analytic classification using a shared frozen pretrained backbone and client-local labeled samples. Its clients contribute information from sample subsets in a common feature space. AFCL~\cite{AFCL} extends analytic fitting to federated continual learning, using frozen features, closed-form local classifiers and recursive global aggregation to accommodate heterogeneous client data and newly arriving classes. Its recursion incorporates sample and class increments within a shared feature space. CALM~\cite{CALM} applies analytic learning to crowdsourced continual service construction. It uses a frozen feature extractor and recursive analytic updates to incorporate incoming worker data and classes. Its continual-update setting differs from VAFL's aligned-record, feature-partitioned setting. These methods demonstrate the utility of frozen representations and analytic classification, providing relevant points of comparison for our head-training approach.

For horizontally partitioned or incrementally arriving samples, local sufficient products use a shared feature space. Vertical partitioning instead requires off-diagonal products between features held by different parties. Thus frozen representations, one-hot targets and analytic fitting are established ingredients, while the vertical setting changes which computations require collaboration. VAFL combines this setting with strategies that reduce the amount and dimension of protected computation, and examines how retaining local predictive information affects the resulting tradeoffs. The theoretical analysis distinguishes exact equivalence from reduced-space approximation, and the experiments assess both predictive quality and fitting cost. This positions VAFL through the scope of its collaborative fitting and acceleration mechanisms while recognizing the analytic properties shared with earlier work.

\FloatBarrier
\endgroup

\section{Primers and Gaps of Analytic Learning: From Horizontal to Vertical Scenarios}
\label{Preliminaries}

As a distinct approach to training NN compared to gradient-based back-propagation, AL has garnered significant attention in recent years.
By directly solving the analytic network's optimal solution without the need for repeated iterations, AL presents an efficient technical alternative to traditional gradient-based methods.
In this section, we aim to present the basic primers and key gaps of AL from horizontal to vertical scenarios.
Specifically, Section~\ref{Horizontal AL} introduces standard analytic updates for sample-partitioned data, before examining the cross-feature products required in the vertical setting.
Subsequently, in Section~\ref{Vertical AL}, we will further discuss and analyze the distinctions and challenges associated with extending AL to vertical scenarios.

\subsection{Methodologies and Principles of Horizontal AL}
\label{Horizontal AL}

Here, we elaborate on the methodologies and underlying principles of current AL-based applications, which are limited to horizontal scenarios.
Fundamentally, the direct application of AL to complex models is infeasible, given the intractability of deriving analytical solutions for their parameters.
Fortunately, this limitation has been addressed with the advent of widely researched pre-trained models.
By leveraging AL to replace the classification heads of pre-trained models, strong performance can be achieved without gradient-related issues.

For clarity, let's first examine standard AL from a centralized perspective, considering a specific optimal weight estimation problem for a regularized linear model as an example, which is the most commonly used method and the one adopted in this paper.
The specific optimization objective is shown below:
\begin{equation}
\label{eq:Section-3-1-1}
    \mathbf{\hat W} = \mathop{\arg\min}\limits_{\mathbf{W}} {\|\mathbf{Y} -\mathbf{\widetilde X} \mathbf{W} \|}_{\text F}^2+\gamma{\| \mathbf{W} \|}_{\text F}^2,
\end{equation}
where $\mathbf{\widetilde X}$ is the feature matrix extracted from the input matrix $\mathbf{X}$ via a pre-trained backbone, $\mathbf{W}$ is the weight matrix, $\mathbf{Y}$ is the target matrix, and $\gamma$ is a regularization parameter that governs the magnitude of the weights. 
The objective is to minimize the error between the predicted outputs $\mathbf{\widetilde X} \mathbf{W}$ and the ground truth $\mathbf{Y}$, while incorporating a regularization term to prevent overfitting.
Since this is a least squares problem with an additional regularization term, its closed-form solution can be easily obtained using least squares estimation techniques (the detailed derivation is provided in Lemma 1 of Appendix~C.1).
The closed-form solution of \eqref{eq:Section-3-1-1} is:
\begin{equation}
\label{eq:Section-3-1-2}
    \mathbf{\hat W} = {({\mathbf{\widetilde X}}^\top\mathbf{\widetilde X}+\gamma\mathbf{I})}^{-1}{\mathbf{\widetilde X}}^\top\mathbf{Y},
\end{equation}
where $\mathbf{I}$ is the identity matrix.
Consequently, AL enables the direct computation of the optimal solution \eqref{eq:Section-3-1-2} via matrix operations, rather than relying on iterative multi-round gradient-based BP. 
In this way, AL avoids dependence on gradients, thus mitigating gradient-related issues such as gradient explosion and vanishing.

Next, we extend our discussion to the HFL scenario to introduce existing horizontal AL methods.
In HFL scenario, \eqref{eq:Section-3-1-1} represents the global optimization objective of existing horizontal AL methods, and \eqref{eq:Section-3-1-2} constitutes the optimal solution accordingly. 
The primary goal of these horizontal AL methods is to decouple the computation of \eqref{eq:Section-3-1-2} across individual parties.
Specifically, in HFL setting, the complete feature matrix $\mathbf{\widetilde X}$ and label matrix $\mathbf{Y}$ can be represented as $[\mathbf{\widetilde X}_{(1)}^\top, \mathbf{\widetilde X}_{(2)}^\top, \dots, \mathbf{\widetilde X}_{(K)}^\top]^\top$ and $[\mathbf{Y}_{(1)}^\top, \mathbf{Y}_{(2)}^\top, \dots, \mathbf{Y}_{(K)}^\top]^\top$, where $\mathbf{\widetilde X}_{(k)}$ and $\mathbf{Y}_{(k)}$ denote the local matrices held by party $k$ in HFL. 
Accordingly, the optimal solution in \eqref{eq:Section-3-1-2} can be reformulated as:
\begin{equation}
\label{eq:Section-3-1-3}
\begin{aligned}
    \mathbf{\hat W} 
    &= ([\mathbf{\widetilde X}_{(1)}^\top, \mathbf{\widetilde X}_{(2)}^\top, \dots, \mathbf{\widetilde X}_{(K)}^\top][\mathbf{\widetilde X}_{(1)}^\top, \mathbf{\widetilde X}_{(2)}^\top, \dots, \mathbf{\widetilde X}_{(K)}^\top]^\top + \gamma \mathbf{I})^{-1}[\mathbf{\widetilde X}_{(1)}^\top, \mathbf{\widetilde X}_{(2)}^\top, \dots, \mathbf{\widetilde X}_{(K)}^\top][\mathbf{Y}_{(1)}^\top, \mathbf{Y}_{(2)}^\top, \dots, \mathbf{Y}_{(K)}^\top]^\top \\
    &= (\sum\nolimits_{k=1}^K\mathbf{\widetilde X}_{(k)}^\top\mathbf{\widetilde X}_{(k)} + \gamma \mathbf{I})^{-1}\sum\nolimits_{k=1}^K\mathbf{\widetilde X}_{(k)}^\top\mathbf{Y}_{(k)}.
\end{aligned}
\end{equation}

It is evident that each party $k$ only needs to upload its local auto-correlation matrices $\mathbf{\widetilde X}_{(k)}^\top\mathbf{\widetilde X}_{(k)}$ and $\mathbf{\widetilde X}_{(k)}^\top\mathbf{Y}_{(k)}$ to facilitate the straightforward computation of the global model according to \eqref{eq:Section-3-1-3}. 
This is precisely the approach adopted by existing horizontal AL methods. 
Nevertheless, applying AL to the VFL scenario proves to be significantly more complex, as detailed in the subsequent section.

\subsection{Distinctions and Challenges of Vertical AL}
\label{Vertical AL}

Here, we provide comprehensive discussions and analyses of the distinctions and challenges associated with extending AL to vertical scenarios.
While existing applications have successfully deployed AL in horizontal settings (sample-level partitioning), where each party need only upload local auto-correlation matrices for additive aggregation, a significant research gap remains regarding the extension of AL to vertical settings (feature-level partitioning).
This disparity stems from the fact that in VFL, the distinct partitioning strategy necessitates the computation of extensive cross-correlation matrices, in addition to auto-correlation matrices.

For clarity, let's first consider the simplest VFL scenario involving only 2 parties.
In this setting, the complete feature matrix $\mathbf{\widetilde X}$ and label matrix $\mathbf{Y}$ can be partitioned as $[\mathbf{\widetilde X}_{1}, \mathbf{\widetilde X}_{2}]$ and $[\mathbf{Y}_{1}, \mathbf{Y}_{2}]$, respectively, where $\{ \mathbf{\widetilde X}_{1}, \mathbf{Y}_{1}\}$ and $\{ \mathbf{\widetilde X}_{2}, \mathbf{Y}_{2}\}$ denote the local matrices held by parties $1$ and $2$, respectively.
By incorporating these partitioned forms $[\mathbf{\widetilde X}_{1}, \mathbf{\widetilde X}_{2}]$ and $[\mathbf{Y}_{1}, \mathbf{Y}_{2}]$ into \eqref{eq:Section-3-1-2}, the optimal global model $\mathbf{\hat W}$ has the closed-form solution:
\begin{equation}
\label{eq:Section-3-2-1}
\begin{aligned}
    \mathbf{\hat W} 
    &= {({\mathbf{\widetilde X}}^\top\mathbf{\widetilde X}+\gamma\mathbf{I})}^{-1}{\mathbf{X}}^\top\mathbf{Y} \\
    &= {([\mathbf{\widetilde X}_{1}, \mathbf{\widetilde X}_{2}]^\top[\mathbf{\widetilde X}_{1}, \mathbf{\widetilde X}_{2}]+\gamma\mathbf{I})}^{-1}[\mathbf{\widetilde X}_{1}, \mathbf{\widetilde X}_{2}]^\top[\mathbf{Y}_{1}, \mathbf{Y}_{2}] \\
    &= \begin{bmatrix} \mathbf{\widetilde X}_1^\top\mathbf{\widetilde X}_1 + \gamma \mathbf{I} & \mathbf{\widetilde X}_1^\top\mathbf{\widetilde X}_2 \\ \mathbf{\widetilde X}_2^\top\mathbf{\widetilde X}_1 & \mathbf{\widetilde X}_2^\top\mathbf{\widetilde X}_2 + \gamma \mathbf{I} \end{bmatrix}^{-1} \begin{bmatrix} \mathbf{\widetilde X}_1^\top\mathbf{Y} \\ \mathbf{\widetilde X}_2^\top\mathbf{Y} \end{bmatrix}.
\end{aligned}
\end{equation}

It is evident that even in this fundamental two-party scenario, the solution for the global model $\mathbf{\hat W}$ in vertical settings becomes significantly more intricate. 
In contrast to horizontal settings, the vertical setting necessitates the computation of cross-correlation matrices between parties, specifically $\mathbf{\widetilde X}_1^\top\mathbf{\widetilde X}_2$ and $\mathbf{\widetilde X}_2^\top\mathbf{\widetilde X}_1$. 
This motivates collaborative computation of inter-party correlations without directly exchanging the original feature blocks.

Subsequently, let's further generalize this analysis to the multi-party VFL scenario involving $K$ parties, which significantly increases the complexity of deriving the global model $\mathbf{\hat W}$.
In this setting, the complete feature matrix $\mathbf{\widetilde X}$ and label matrix $\mathbf{Y}$ can be partitioned as $[\mathbf{\widetilde X}_{1}, \mathbf{\widetilde X}_{2}, \dots, \mathbf{\widetilde X}_{K}]$ and $[\mathbf{Y}_{1}, \mathbf{Y}_{2}, \dots, \mathbf{Y}_{K}]$, respectively. 
Accordingly, the corresponding closed-form solution for the optimal global model $\mathbf{\hat W}$ can be derived as follows:
\begin{equation}
\label{eq:Section-3-2-2}
\begin{aligned}
    \mathbf{\hat W} 
    &= {({\mathbf{\widetilde X}}^\top\mathbf{\widetilde X}+\gamma\mathbf{I})}^{-1}{\mathbf{X}}^\top\mathbf{Y} \\
    &= {([\mathbf{\widetilde X}_{1}, \mathbf{\widetilde X}_{2}, \dots, \mathbf{\widetilde X}_{K}]^\top[\mathbf{\widetilde X}_{1}, \mathbf{\widetilde X}_{2}, \dots, \mathbf{\widetilde X}_{K}]+\gamma\mathbf{I})}^{-1}[\mathbf{\widetilde X}_{1}, \mathbf{\widetilde X}_{2}, \dots, \mathbf{\widetilde X}_{K}]^\top[\mathbf{Y}_{1}, \mathbf{Y}_{2}, \dots, \mathbf{Y}_{K}] \\
    &= \begin{bmatrix} \mathbf{\widetilde X}_1^\top\mathbf{\widetilde X}_1 + \gamma \mathbf{I} & \mathbf{\widetilde X}_1^\top\mathbf{\widetilde X}_2 & \cdots & \mathbf{\widetilde X}_1^\top\mathbf{\widetilde X}_K \\ \mathbf{\widetilde X}_2^\top\mathbf{\widetilde X}_1 & \mathbf{\widetilde X}_2^\top\mathbf{\widetilde X}_2 + \gamma \mathbf{I} & \cdots & \mathbf{\widetilde X}_2^\top\mathbf{\widetilde X}_K \\
    \vdots & \vdots & \ddots & \vdots \\
    \mathbf{\widetilde X}_K^\top\mathbf{\widetilde X}_1 & \mathbf{\widetilde X}_K^\top\mathbf{\widetilde X}_2 & \cdots & \mathbf{\widetilde X}_K^\top\mathbf{\widetilde X}_K + \gamma \mathbf{I}    \end{bmatrix}^{-1} \begin{bmatrix} \mathbf{\widetilde X}_1^\top\mathbf{Y} \\ \mathbf{\widetilde X}_2^\top\mathbf{Y} \\ \vdots \\  \mathbf{\widetilde X}_K^\top\mathbf{Y} \end{bmatrix}.
\end{aligned}
\end{equation}

It is evident that in the multi-party VFL scenario, the quantity of cross-correlation matrices required for computation escalates drastically, further exacerbating the complexity of deriving the global model $\mathbf{\hat W}$.
Specifically, with a total of $K$ participating parties, the number of required cross-correlation matrices scales to $K^2 - K$.
Consequently, the necessity of computing numerous inter-party cross-correlation matrices without compromising data privacy is the main technical challenge in extending AL from HFL to VFL.

Moreover, this transition also introduces additional technical challenges.
First, unlike HFL, which constructs a single shared global model, VFL requires the derivation of multiple partitioned local models.
Consequently, a federated implementation must specify the messages and model blocks assigned to each party, together with the information available in its intermediate computations.
Second, given that each party in VFL focuses on distinct feature spaces, the utilization of a unified pre-trained backbone for feature extraction is infeasible. 
This requires flexible backbone selection for each party.

\begin{figure}
    \centering
    \includegraphics[width=1.0\linewidth]{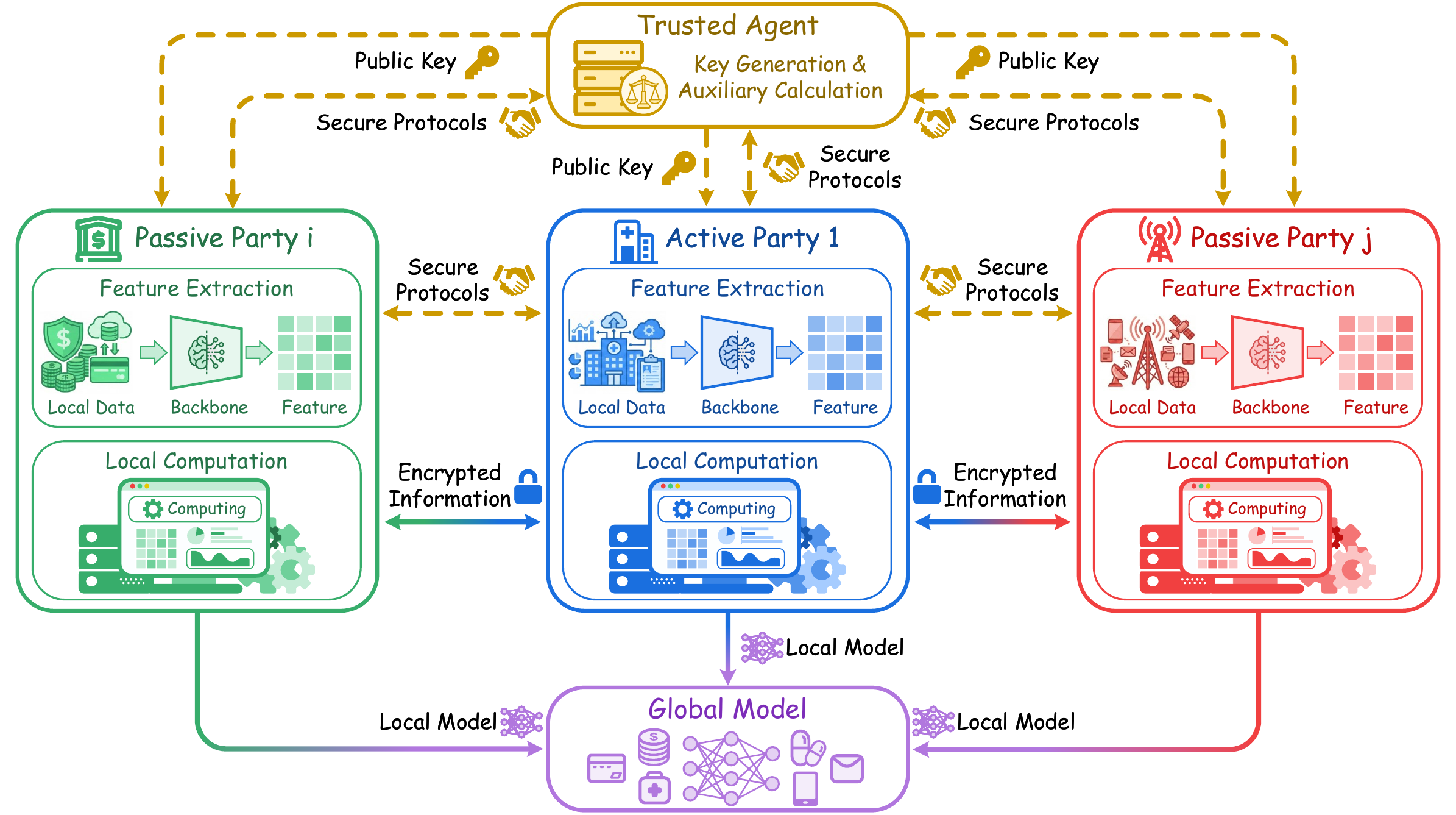}
    \caption{System model of VAFL.}
    \label{fig:fedav-overview}
\end{figure}

\section{Our Proposed VAFL}
In this section, we present VAFL, which extends analytic learning to vertically partitioned data through recursive least squares and secure matrix computation.

\noindent\textbf{VFL setting and learning objective.}
We consider $K$ parties whose records have been aligned to the same $\text N$ samples before training.
Party $1$ is the active party and holds local records $\mathbf{X}_1\in\mathbb{R}^{\text N\times\text F_1}$ together with the target matrix $\mathbf{Y}\in\mathbb{R}^{\text N\times\text C}$.
Each passive party $k\in\{2,\ldots,K\}$ holds only its local records $\mathbf{X}_k\in\mathbb{R}^{\text N\times\text F_k}$.
Here, $\text F_k$ denotes the number of features held by party $k$, and $\text C$ is the output dimension, corresponding to the number of one-hot encoded classes for classification or one output for scalar regression.
The parties retain their raw records locally and collaborate through protected matrix exchanges.
Each party supplies a fixed local representation $\mathbf{\widetilde X}_k\in\mathbb{R}^{\text N\times\widetilde{\text F}_k}$, as defined below, and the corresponding centralized representation is $\mathbf{\widetilde X}=[\mathbf{\widetilde X}_1,\ldots,\mathbf{\widetilde X}_K]$.
Our objective is to obtain the local model blocks $\{\mathbf{\hat W}_k\}_{k=1}^K$ without pooling the parties' raw features or labels.
For the full-dimensional method, their vertical concatenation $\mathbf{\hat W}=[\mathbf{\hat W}_1^\top,\ldots,\mathbf{\hat W}_K^\top]^\top$ must equal the unique centralized solution of \eqref{eq:Section-3-1-1} for the same representations, targets and regularization parameter $\gamma>0$, in exact arithmetic.
This formulation preserves the centralized analytic-learning solution while distributing its computation and local model blocks according to the local feature partition of each participating party.

\noindent\textbf{Privacy objective and threat model.}
Our privacy objective concerns each honest party's raw input records $\mathbf X_k$ and local feature matrix $\mathbf{\widetilde X}_k$ in its original feature coordinates.
We seek to limit their disclosure and reconstruction during collaborative fitting while keeping raw inputs and local representations at their owners.
Target-label confidentiality and secrecy of the prescribed model outputs are outside this objective. Labels and outputs available to an observer are nevertheless included in its auxiliary information when analyzing feature reconstruction.
The auxiliary agent holds the Paillier secret key and performs the prescribed decryptions and masked plaintext computations, while the parties hold the public key and retain their private local transformations.

\textit{(1) Non-colluding semi-honest model.}
Following prior VFL studies~\cite{semi-3,semi-11}, parties and the agent execute the protocols without collusion but may inspect their received messages, randomness and outputs.
Each role observes its own inputs and randomness, public dimensions and encoding metadata, messages addressed to it, and its assigned outputs.
The analysis distinguishes ciphertext confidentiality against observers without the secret key from feature reconstruction using the agent's plaintext observations.

\textit{(2) Colluding participants.}
We examine parties that pool their observations while the agent remains non-colluding, and coalitions that also include the agent and its secret key.
These cases assume the prescribed message generation and randomness rules.
The reconstruction target is an honest party's feature data or raw input records, and that party retains its private local transformation.
Section~\ref{subsec:Privacy Analyses} establishes the ciphertext and coordinate-transformation properties supporting this objective.
Appendix~D gives the conditional reconstruction arguments, the information exposed to each role and coalition, and the limits of these properties. Arbitrary active deviations are outside the threat model.

We first derive the recursive least-squares formulation, describe its secure federated implementation in Section~\ref{Sec:Federated Implementation}, and introduce acceleration strategies in Section~\ref{sec:acceleration}. Section~\ref{Sec:Theoretical-Analyses} analyzes validity, privacy properties and efficiency.

\subsection{Vertically Recursive Least Squares}

To better illustrate the motivation and insights behind our proposed VAFL, we first describe it from a centralized perspective. The detailed federated implementation of VAFL will be introduced in the subsequent Section~\ref{Sec:Federated Implementation}.

Specifically, leveraging the exceptional performance and widespread applicability of pre-trained backbones, we employ them to extract features from the local samples of each party $k$, as formulated below:
\begin{equation}
\label{eq:section-4-1-1}
    \mathbf{\widetilde{X}}_k  = \mathrm{Flatten}({\mathrm{Backbone}_k}(\mathbf{X}_k, \bm{\Theta}_k)),
\end{equation}
where $\mathbf{\widetilde{X}}_k \in \mathbb{R}^{\text N \times \widetilde{\text F}_k}$ denotes the extracted features of the party $k$ with dimension $\widetilde{\text F}_k$, while $\mathrm{Backbone}_k$ represents its pre-trained backbone with frozen parameters $\bm{\Theta}_k$.
Notably, our VAFL exhibits exceptional versatility, allowing each party to select an appropriate backbone based on the nature of its samples and available computational resources, ranging from heavyweight pre-trained models to simple operations like the flatten function.

Subsequently, building upon the extracted features $\{ \mathbf{\widetilde X}_k \}_{k=1}^K$ and the labels $\mathbf{Y}$, we proceed to construct the optimal local models $\{ \mathbf{\hat W}_k \}_{k=1}^K$. 
Notably, as analyzed in Section~\ref{Preliminaries}, these optimal local models $\{ \mathbf{\hat W}_k \}_{k=1}^K$ are essentially partitions of the optimal complete model $\mathbf{\hat W}$, whose analytical expression is provided in \eqref{eq:Section-3-1-2}.

First, for the sake of clarity, let's consider the two-party case (i.e., $K=2$).
By substituting the complete feature matrix $\mathbf{\widetilde X}$ in \eqref{eq:Section-3-1-2} with the partitioned form $[\mathbf{\widetilde X}_{1}, \mathbf{\widetilde X}_{2}]$, we can derive the corresponding optimization solution:
\begin{equation}
\label{eq:section-4-1-3}
    \begin{bmatrix} \mathbf{\hat W}_1 \\ \mathbf{\hat W}_2 \end{bmatrix} 
    = { \left({\begin{bmatrix} \mathbf{\widetilde X}_{1}^\top \\ \mathbf{\widetilde X}_{2}^\top \end{bmatrix}}[\mathbf{\widetilde X}_{1}, \mathbf{\widetilde X}_{2}]+\gamma\mathbf{I} \right)}^{-1}\begin{bmatrix} \mathbf{\widetilde X}_{1}^\top \\ \mathbf{\widetilde X}_{2}^\top \end{bmatrix}\mathbf{Y}
    = \begin{bmatrix} \mathbf{\widetilde X}_1^\top\mathbf{\widetilde X}_1 + \gamma \mathbf{I} & \mathbf{\widetilde X}_1^\top\mathbf{\widetilde X}_2 \\ \mathbf{\widetilde X}_2^\top\mathbf{\widetilde X}_1 & \mathbf{\widetilde X}_2^\top\mathbf{\widetilde X}_2 + \gamma \mathbf{I} \end{bmatrix}^{-1} \begin{bmatrix} \mathbf{\widetilde X}_1^\top\mathbf{Y} \\ \mathbf{\widetilde X}_2^\top\mathbf{Y} \end{bmatrix}.
\end{equation}

For notational convenience, we define the \textit{Feature Gram Matrix} $\mathbf{F}_{i,j}$ and \textit{Label Gram Matrix} $\mathbf{L}_i$ as follows:
\begin{equation}
\label{eq:section-4-1-4}
    \mathbf{F}_{i,j} = \mathbf{\widetilde X}_i^\top\mathbf{\widetilde X}_j, \quad \mathbf{L}_i=\mathbf{\widetilde X}_i^\top\mathbf{Y}.
\end{equation}

Leveraging the matrix block inversion theorem, we can further simplify \eqref{eq:section-4-1-3}, thereby deriving the explicit expressions for the local models as follows (the detailed proof is provided in Theorem 1 of Section~\ref{Sec:Theoretical-Analyses}):
\begin{equation}
\label{eq:section-4-1-5}
    \begin{cases}
        \mathbf{\hat W}_1 = \mathbf{U}_1\mathbf{L}_1 - (\mathbf{F}_{1,1} + \gamma\mathbf{I})^{-1} \mathbf{F}_{1,2} \mathbf{U}_2\mathbf{L}_2, \\
        \mathbf{\hat W}_2 = \mathbf{U}_2\mathbf{L}_2 - (\mathbf{F}_{2,2} + \gamma\mathbf{I})^{-1} \mathbf{F}_{2,1} \mathbf{U}_1\mathbf{L}_1,
    \end{cases}
\end{equation}
where
\begin{equation}
\label{eq:section-4-1-6}
    \begin{cases}
        \mathbf{U}_1 = [(\mathbf{F}_{1,1} + \gamma\mathbf{I}) - \mathbf{F}_{1,2} (\mathbf{F}_{2,2} + \gamma\mathbf{I})^{-1} \mathbf{F}_{2,1}]^{-1},\\
        \mathbf{U}_2 = [(\mathbf{F}_{2,2} + \gamma\mathbf{I}) - \mathbf{F}_{2,1} (\mathbf{F}_{1,1} + \gamma\mathbf{I})^{-1} \mathbf{F}_{1,2}]^{-1},  \\
    \end{cases}
\end{equation}

Second, we extend our analysis to the more intricate multi-party case.
By observing \eqref{eq:Section-3-1-2} and \eqref{eq:Section-3-2-2}, we note that the primary challenge lies in computing the inverse term $({\mathbf{\widetilde X}}^\top\mathbf{\widetilde X}+\gamma\mathbf{I})^{-1}$, which involves substantial auto-correlation and cross-correlation matrices.
Define the \textit{Aggregated Gram Inverse} $\mathbf{R}_k$ as:
\begin{equation}
\label{eq:section-3-1-7}
    \mathbf{R}_k = (\mathbf{\widetilde X}_{1:k}^\top\mathbf{\widetilde X}_{1:k}+\gamma\mathbf{I})^{-1} = (\mathbf{F}_{1:k,1:k}+\gamma\mathbf{I})^{-1}.
\end{equation}

Then, we focus on the recursive computation of the \textit{Aggregated Gram Inverse} $\mathbf{R}_k$. 
Specifically, by expanding $\mathbf{\widetilde X}_{1:k}$ in \eqref{eq:section-3-1-7} as $[\mathbf{\widetilde X}_{1:k-1}, \mathbf{\widetilde X}_{k}]$, we obtain the following recursive formula:
\begin{equation}
\label{eq:section-3-1-8}
    \mathbf{R}_k = \begin{bmatrix} \mathbf{\widetilde X}_{1:k-1}^\top\mathbf{\widetilde X}_{1:k-1} + \gamma \mathbf{I} & \mathbf{\widetilde X}_{1:k-1}^\top\mathbf{\widetilde X}_{k} \\ \mathbf{\widetilde X}_{k}^\top\mathbf{\widetilde X}_{1:k-1} & \mathbf{\widetilde X}_{k}^\top\mathbf{\widetilde X}_{k} + \gamma \mathbf{I} \end{bmatrix}^{-1}
    = \begin{bmatrix} \mathbf{R}_{k-1}^{-1} & \mathbf{F}_{1:k-1,k} \\ \mathbf{F}_{k,1:k-1} & \mathbf{F}_{k,k} + \gamma \mathbf{I} \end{bmatrix}^{-1}.
\end{equation}

Leveraging the matrix block inversion theorem, we can further simplify \eqref{eq:section-3-1-8} to eliminate the inverse operation, thereby deriving the explicit recursive formulas of \textit{Aggregated Gram Inverse} as follows (the detailed proof for all four recursive blocks is provided in Theorem 2 of Section~\ref{Sec:Theoretical-Analyses}):
\begin{equation}
\label{eq:section-4-1-9}
\mathbf{R}_k = 
\begin{bmatrix}
    \mathbf{R}_k^{(1)} & \mathbf{R}_k^{(2)} \\
    \mathbf{R}_k^{(3)} & \mathbf{R}_k^{(4)}
\end{bmatrix},
\end{equation}
where
\begin{equation}
\label{eq:section-4-1-10}
\begin{cases}
\mathbf{R}^{(1)}_k = \mathbf{R}_{k-1} + \mathbf{R}_{k-1} \mathbf{F}_{1:k-1,k} \mathbf{R}^{(4)}_k \mathbf{F}_{k,1:k-1} \mathbf{R}_{k-1}, \\
\mathbf{R}^{(2)}_k = -\mathbf{R}_{k-1}\mathbf{F}_{1:k-1,k}\mathbf{R}^{(4)}_k = \mathbf{R}^{(3)\top}_k,\\
\mathbf{R}^{(3)}_k = -(\mathbf{F}_{k,k} + \gamma \mathbf{I})^{-1} \mathbf{F}_{k,1:k-1} \mathbf{R}^{(1)}_k = \mathbf{R}^{(2)\top}_k,\\
\mathbf{R}^{(4)}_k = ( \mathbf{F}_{k,k} + \gamma \mathbf{I} - \mathbf{F}_{k,1:k-1} \mathbf{R}_{k-1} \mathbf{F}_{1:k-1,k} )^{-1}.
\end{cases}
\end{equation}

By applying the vertical recursive formulas \eqref{eq:section-4-1-9} and \eqref{eq:section-4-1-10} across all parties, we derive the final \textit{Aggregated Gram Inverse} $\mathbf{R}_K$ that aggregates all local information.
Finally, the complete model $\mathbf{\hat W}$ can be computed according to \eqref{eq:section-4-1-11}, and each party's local model can then be obtained by partitioning the complete model $\mathbf{\hat W}$ accordingly.
\begin{equation}
\label{eq:section-4-1-11}
    \mathbf{\hat W} = \mathbf{R}_K \mathbf{L}_{1:K} = [\mathbf{\hat W}_{1}^\top, \mathbf{\hat W}_{2}^\top, \cdots, \mathbf{\hat W}_{K}^\top]^\top , \quad \mathbf{L}_{1:K} = [\mathbf{L}_{1}^\top, \mathbf{L}_{2}^\top, \cdots, \mathbf{L}_{K}^\top]^\top.
\end{equation}

\subsection{Federated Implementation with Secure Protocols}
\label{Sec:Federated Implementation}

Here, we elaborate on the federated implementation of our VAFL.
Specifically, we first describe three computation protocols using additive homomorphic encryption (AHE), local perturbations and additive shares.
Building upon these protocols, we then present the specific federated implementations of VAFL for both the two-party and multi-party cases.
The detailed procedures are provided as follows.

First of all, to ensure generality, we adopt the widely used Paillier AHE scheme~\cite{homomorphic-Paillier} as the cryptographic basis for encrypted messages, noting that other suitable AHE schemes can also be employed in our VAFL.
For conciseness, we outline the primary properties of the Paillier scheme here, and detailed preliminaries are provided in Appendix~A.
Specifically, the Paillier scheme supports homomorphic addition and subtraction between ciphertexts, as well as homomorphic multiplication between a ciphertext and a plaintext.
Let $\llbracket \cdot \rrbracket$ denote the encryption operation, and let $\oplus$, $\ominus$, and $\otimes$ represent the homomorphic addition, subtraction, and multiplication operators, respectively. 
For any arbitrary matrices $\mathbf{A}$, $\mathbf{B}$, and $\mathbf{C}$, the following properties hold:
\begin{equation}
    \label{q:section-4-2-1}
     \llbracket \mathbf{A} \rrbracket \oplus \llbracket \mathbf{B} \rrbracket = \llbracket \mathbf{A+B} \rrbracket, \quad \llbracket \mathbf{A} \rrbracket \ominus \llbracket \mathbf{B} \rrbracket = \llbracket \mathbf{A-B} \rrbracket, \quad \llbracket \mathbf{B} \rrbracket \otimes \mathbf{C} = \mathbf{B} \otimes \llbracket \mathbf{C} \rrbracket  = \llbracket \mathbf{B}\mathbf{C} \rrbracket.
\end{equation}

Subsequently, by observing the centralized formulations for both the two-party case (i.e., \eqref{eq:section-4-1-5} and \eqref{eq:section-4-1-6}) and the multi-party case (i.e., \eqref{eq:section-4-1-9} and \eqref{eq:section-4-1-10}), it becomes evident that the implementation of VAFL necessitates matrix multiplication and matrix inversion operations.
Nonetheless, these operations are not supported by the AHE scheme.
To address this, we introduce Protocol 1 for secure matrix multiplication and Protocol 2 for secure matrix inversion, both of which leverage an auxiliary trusted agent.
The agent follows the prescribed computations and holds the decryption key. Raw records and labels are not uploaded directly, while the decrypted masked requests can reveal derived information as detailed in Section~\ref{subsec:Privacy Analyses}.
Furthermore, given that VAFL requires the computation of numerous cross-correlation gram matrices, we devise Protocol 3 specifically for this purpose. 
The detailed descriptions and workflows of these protocols are as follows.

\noindent\textbf{Protocol 1 (Secure Matrix Multiplication):} 
This protocol uses AHE and an auxiliary agent to compute an encrypted matrix product through additively masked operands.
Specifically, for any party $k$ holding ciphertexts $\llbracket \mathbf A \rrbracket $ and $\llbracket \mathbf B \rrbracket $ (where $\mathbf{A} \in \mathbb{R}^{n \times p}$ and $\mathbf{B} \in \mathbb{R}^{p \times m}$), the protocol computes $\llbracket \mathbf{A} \mathbf{B} \rrbracket$ with the aid of the trusted agent possessing the decryption key.
The agent observes the additively perturbed operands and computes their product. Confidentiality of encrypted components and the information present in these plaintext observations are analyzed separately in Section~\ref{subsec:Privacy Analyses}. 
The party removes the mask terms homomorphically to obtain the encrypted product.
We denote this protocol by $\text{SMM}(\cdot,\cdot)$ in \eqref{eq:Protocol-1-1}, with the steps detailed below:
\begin{equation}
\label{eq:Protocol-1-1}
    \text{SMM}(\llbracket \mathbf A \rrbracket, \llbracket \mathbf B \rrbracket) = \llbracket \mathbf{AB} \rrbracket.
\end{equation}
\begin{itemize}
    \item \textbf{Step 1:} 
    Party $k$ generates two random perturbation matrices $\mathbf{P}_\text{A} \in \mathbb{R}^{n \times p}$ and $\mathbf{P}_\text{B} \in \mathbb{R}^{p \times m}$, and encrypts them using the public key to obtain $\llbracket \mathbf{P}_\text{A} \rrbracket$ and $\llbracket \mathbf{P}_\text{B} \rrbracket$. 
    Subsequently, party $k$ locally perturbs the ciphertexts $\llbracket \mathbf A \rrbracket$ and $\llbracket \mathbf B \rrbracket$ by computing $\llbracket \mathbf{\widetilde A} \rrbracket = \llbracket \mathbf A \rrbracket \oplus \llbracket \mathbf{P}_\text{A} \rrbracket$ and $\llbracket \mathbf{\widetilde B} \rrbracket = \llbracket \mathbf B \rrbracket \oplus \llbracket \mathbf{P}_\text{B} \rrbracket$. 
    Finally, party $k$ transmits the perturbed ciphertexts $\llbracket \mathbf{\widetilde A} \rrbracket$ and $\llbracket \mathbf{\widetilde B} \rrbracket$ to the trusted agent.

    \item \textbf{Step 2:}
    The agent decrypts $\llbracket \mathbf{\widetilde A} \rrbracket$ and $\llbracket \mathbf{\widetilde B} \rrbracket$ to obtain the plaintexts $\mathbf{\widetilde A}$ and $\mathbf{\widetilde B}$. 
    Subsequently, the agent computes the product $\mathbf{\widetilde A}\mathbf{\widetilde B}$, re-encrypts the perturbed result to obtain $\llbracket \mathbf{\widetilde A}\mathbf{\widetilde B} \rrbracket$, and returns it to party $k$.
    
    \item \textbf{Step 3:}
    Party $k$ derives the desired $\llbracket \mathbf{A} \mathbf{B} \rrbracket$ according to \eqref{eq:protocol-1-1}, by leveraging the received $\llbracket \mathbf{\widetilde A}\mathbf{\widetilde B} \rrbracket$ alongside locally stored ciphertexts $\llbracket \mathbf{A} \rrbracket$ and $\llbracket \mathbf{B} \rrbracket$, as well as perturbation matrices $\mathbf{P}_\text{A}$ and $\mathbf{P}_\text{B}$.
    \begin{equation}
    \label{eq:protocol-1-1}
        \llbracket \mathbf{A} \mathbf{B} \rrbracket = \llbracket \mathbf{\widetilde A}\mathbf{\widetilde B} \rrbracket \ominus (\llbracket \mathbf{A} \rrbracket \otimes  \mathbf{P}_\text{B} ) \ominus (\mathbf{P}_\text{A} \otimes \llbracket \mathbf{B} \rrbracket) \ominus \llbracket \mathbf{P}_\text{A}\mathbf{P}_\text{B} \rrbracket.
    \end{equation}
\end{itemize}

\noindent\textbf{Protocol 2 (Secure Matrix Inversion):} 
This protocol uses AHE and an auxiliary agent to compute an encrypted matrix inverse through a multiplicatively masked request. Its algebraic recovery identity and agent-visible information are distinguished in Section~\ref{subsec:Privacy Analyses}.
Specifically, any party $k$ holding a ciphertext $\llbracket \mathbf A \rrbracket$ of an invertible matrix $\mathbf A$ can compute the encrypted inverse $\llbracket \mathbf A^{-1} \rrbracket$ with the aid of the trusted agent who possesses the private key required for decryption.
We denote this protocol by $\text{SMI}(\cdot)$ in \eqref{eq:Protocol-2-1}, with the steps detailed below:
\begin{equation}
\label{eq:Protocol-2-1}
    \text{SMI}(\llbracket \mathbf A \rrbracket ) = \llbracket \mathbf{A}^{-1} \rrbracket.
\end{equation}
\begin{itemize}
    \item \textbf{Step 1:}
    Party $k$ generates an invertible perturbation matrix $\mathbf{P}_\text{A} \in \mathbb{R}^{n \times n}$ and encrypts $\mathbf A$ by computing $\llbracket \mathbf{\widetilde A} \rrbracket =\mathbf{P}_\text{A} \otimes \llbracket \mathbf A \rrbracket$. 
    Subsequently, the party transmits the perturbed ciphertext $\llbracket \mathbf{\widetilde A} \rrbracket$ to the agent.
    \item \textbf{Step 2:}
    The agent decrypts the perturbed ciphertext $\llbracket \mathbf{\widetilde A} \rrbracket$ to obtain $\mathbf{\widetilde A}$, computes its inverse $\mathbf{\widetilde A}^{-1}$, and transmits the perturbed inverse $\llbracket \mathbf{\widetilde A}^{-1} \rrbracket$ back to party $k$. 
    \item \textbf{Step 3:}
    Party $k$ utilizes the received $\llbracket \mathbf{\widetilde A}^{-1} \rrbracket$ along with the invertible perturbation matrix $\mathbf{P}_\text{A}$ to derive the desired $\llbracket \mathbf{A}^{-1} \rrbracket$ by computing $\llbracket \mathbf{A}^{-1} \rrbracket = \llbracket \mathbf{\widetilde A}^{-1} \rrbracket \otimes \mathbf{P}_\text{A}$.
\end{itemize}

\noindent\textbf{Protocol 3 (Secure Gram Matrix Computation):} 
This protocol enables two parties, $i$ and $j$, holding private matrices $\mathbf{A}_i \in \mathbb{R}^{p \times n}$ and $\mathbf{{B}}_j \in \mathbb{R}^{p \times m}$ respectively, to collaboratively compute a perturbed product $\mathbf{C}_{i,j}^{\text{Enc}}$.
Each party obtains an additive share of the perturbed product $\mathbf{C}_{i,j}^{\text{Enc}}$, denoted as $\mathbf{C}_{i,j}^{(i)}$ and $\mathbf{C}_{i,j}^{(j)}$, such that:
\begin{equation}
\label{eq:protocol-3-1}
\mathbf{C}_{i,j}^{\text{Enc}}=\mathbf{C}_{i,j}^{(i)}+\mathbf{C}_{i,j}^{(j)} = \bm{\Pi}_i^\top \mathbf{{A}}_i^\top  \mathbf{{B}}_j \bm{\Pi}_j,
\end{equation}
where $\bm{\Pi}_i \in \mathbb{R}^{n \times n'}$ and $\bm{\Pi}_j \in \mathbb{R}^{m \times m'}$ are local coordinate transformations generated by parties $i$ and $j$. Their algebraic effect is distinct from cryptographic encryption.
The exact model-recovery relations in Theorems 3 and 4 of Section~\ref{subsec:Validity Analysis} apply to square orthogonal feature transformations and invertible label transformations, respectively.
This protocol proceeds as follows:
\begin{itemize}
    \item \textbf{Step 1:} Parties $i$ and $j$ locally generate random matrices $\bm{\Pi}_i \in \mathbb{R}^{n \times n'}$ and $\bm{\Pi}_j \in \mathbb{R}^{m \times m'}$. 
    Subsequently, they send their matrix dimensions $p \times n'$ and $p \times m'$ to the agent.
    
    \item \textbf{Step 2:} 
    Upon receiving the dimensions, the agent generates random perturbation matrices $\mathbf{P}_i \in \mathbb{R}^{p \times n'}$ and $\mathbf{P}_j \in \mathbb{R}^{p \times m'}$. 
    It then computes $\mathbf{P}_i^\top \mathbf{P}_j$ and splits it into two additive shares $\mathbf{D}_{i,j}^{(i)}$ and $\mathbf{D}_{i,j}^{(j)}$ satisfying $\mathbf{P}_i^\top \mathbf{P}_j
    = \mathbf{D}_{i,j}^{(i)} + \mathbf{D}_{i,j}^{(j)}$.
    Subsequently, the agent sends $\{ \mathbf{P}_i, \mathbf{D}_{i,j}^{(i)} \}$ to party $i$ and $\{ \mathbf{P}_j, \mathbf{D}_{i,j}^{(j)} \}$ to party $j$.
    
    \item \textbf{Step 3:} 
    Parties $i$ and $j$ locally compute $(\mathbf{{A}}_i\bm{\Pi}_i-\mathbf{P}_i)$ and $(\mathbf{{B}}_j\bm{\Pi}_j-\mathbf{P}_j)$, and exchange these results.

    \item \textbf{Step 4:} 
    Parties $i$ and $j$ compute their additive shares of the permuted product $\mathbf{C}_{i,j}^{(i)}$ and $\mathbf{C}_{i,j}^{(j)}$ as follows:
    \begin{equation}
    \label{eq:protocol-3-2}
    \mathbf{C}_{i,j}^{(i)} = \mathbf{D}_{i,j}^i + \bm{\Pi}_i^\top\mathbf{{A}}_i^\top(\mathbf{{B}}_j\bm{\Pi}_j-\mathbf{P}_j),
    \end{equation}
    \begin{equation}
    \label{eq:protocol-3-3}
    \mathbf{C}_{i,j}^{(j)} = \mathbf{D}_{i,j}^j + (\mathbf{{A}}_i\bm{\Pi}_i-\mathbf{P}_i)^\top \mathbf{{B}}_j \bm{\Pi}_j - (\mathbf{{A}}_i\bm{\Pi}_i-\mathbf{P}_i)^\top(\mathbf{{B}}_j\bm{\Pi}_j-\mathbf{P}_j).
    \end{equation}
    As demonstrated in \eqref{eq:protocol-3-4}, summing the resulting $\mathbf{C}_{i,j}^{(i)}$ and $\mathbf{C}_{i,j}^{(j)}$ can recover the target product $\bm{\Pi}_i^\top\mathbf{{A}}_i^\top\mathbf{{B}}_j\bm{\Pi}_j$:
    \begin{equation}
    \label{eq:protocol-3-4}
    \begin{aligned}
        \mathbf{C}_{i,j}^{(i)} + \mathbf{C}_{i,j}^{(j)} &= \mathbf{D}_{i,j}^i + \bm{\Pi}_i^\top\mathbf{{A}}_i^\top(\mathbf{{B}}_j\bm{\Pi}_j-\mathbf{P}_j) + \mathbf{D}_{i,j}^j +(\mathbf{{A}}_i\bm{\Pi}_i-\mathbf{P}_i)^\top \mathbf{{B}}_j \bm{\Pi}_j - (\mathbf{ {A}}_i\bm{\Pi}_i-\mathbf{P}_i)^\top(\mathbf{ {B}}_j\bm{\Pi}_j-\mathbf{P}_j) \\
        &= \mathbf{P}_i^\top \mathbf{P}_j + \bm{\Pi}_i^\top\mathbf{ {A}}_i^\top(\mathbf{ {B}}_j\bm{\Pi}_j-\mathbf{P}_j) + (\bm{\Pi}_i^\top\mathbf{ {A}}_i^\top-\mathbf{P}_i^\top) \mathbf{ {B}}_j \bm{\Pi}_j - (\bm{\Pi}_i^\top\mathbf{ {A}}_i^\top-\mathbf{P}_i^\top)(\mathbf{ {B}}_j\bm{\Pi}_j-\mathbf{P}_j) \\
        &= \mathbf{P}_i^\top \mathbf{P}_j + \bm{\Pi}_i^\top\mathbf{ {A}}_i^\top(\mathbf{ {B}}_j\bm{\Pi}_j-\mathbf{P}_j) + (\bm{\Pi}_i^\top\mathbf{ {A}}_i^\top-\mathbf{P}_i^\top)\mathbf{P}_j \\
        &= \mathbf{P}_i^\top \mathbf{P}_j + \bm{\Pi}_i^\top\mathbf{ {A}}_i^\top\mathbf{ {B}}_j\bm{\Pi}_j - \bm{\Pi}_i^\top\mathbf{ {A}}_i^\top\mathbf{P}_j + \bm{\Pi}_i^\top\mathbf{ {A}}_i^\top\mathbf{P}_j - \mathbf{P}_i^\top\mathbf{P}_j = \bm{\Pi}_i^\top\mathbf{ {A}}_i^\top\mathbf{ {B}}_j\bm{\Pi}_j.
    \end{aligned}
    \end{equation}
\end{itemize}

We compose Protocols 1--3 to implement the recursive least-squares solution on distributed feature blocks. Each party computes its feature auto-correlation locally, while Protocol 3 constructs encrypted cross-party feature and feature--target products under local coordinate transformations. Protocols 1 and 2 supply the matrix multiplications and inverse requests needed for the model computation.

For $K=2$, the active party combines the local inverses and cross-party products to evaluate the two Schur complements and obtain the encrypted local model blocks. Appendix~B.1 gives the five-step workflow and its equations, with the complete pseudocode in Appendix Algorithm~1.

For $K>2$, the parties successively update the encrypted aggregated Gram inverse as each feature block is incorporated. The last party returns the inverse to the active party, which combines it with the feature--target products, removes the label-coordinate transformation, and distributes the encrypted model blocks. Appendix~B.2 gives the corresponding steps and equations, summarized in Appendix Algorithm~2.

Each party recovers its assigned model block through masked decryption and removal of its retained feature transformation. For the same supplied representations, targets and regularization, the concatenated blocks equal the centralized solution in exact arithmetic, as established in Section~\ref{subsec:Validity Analysis}. Section~\ref{subsec:Privacy Analyses} characterizes the information visible to each party and the agent during this workflow.

\subsection{Acceleration Strategies for VAFL}
\label{sec:acceleration}

Wide feature blocks can make encrypted matrix computation costly. We consider an exact solver optimization and two alternative approximation designs. Below, $\mathbf{X}_k$ denotes the fixed representation $\widetilde{\mathbf{X}}_k$, including any retained intercept, and $\mathbf{X}=[\mathbf{X}_1,\ldots,\mathbf{X}_K]$.

\noindent\textbf{Exact full-dimensional solving (VAFL-E).}
VAFL-E uses block factorization and forward/backward substitution to obtain the model directly, avoiding explicit assembly of the complete Gram inverse. It preserves the original Ridge solution in exact arithmetic, although the dense factorization remains cubic. Its benefit depends on feature and output widths.

\noindent\textbf{Local projection (RP/SP).}
Each party constructs an orthonormal map $\mathbf{Q}_k$: random projection (RP) uses a Gaussian-QR basis, while spectral projection (SP) retains leading eigenvectors of its training-only local feature Gram matrix. With $\mathbf{Q}=\operatorname{diag}(\mathbf{Q}_1,\ldots,\mathbf{Q}_K)$ and $\mathbf{Z}=\mathbf{X}\mathbf{Q}$, both methods compute
\begin{equation}
\mathbf{U}=(\mathbf{Z}^\top \mathbf{Z}+\gamma \mathbf{I})^{-1}\mathbf{Z}^\top \mathbf{Y},\qquad \mathbf{W}^{\mathrm{proj}}=\mathbf{Q}\mathbf{U}.
\label{eq:sp-lift}
\end{equation}
Lifting preserves the projected predictor, $\mathbf{X}\mathbf{W}^{\mathrm{proj}}=\mathbf{Z}\mathbf{U}$, but omitted directions can affect accuracy. The intercept is retained separately. RP serves as a projection control.

\noindent\textbf{Full-local fitting with low-rank correction (LC).}
LC preserves each full-dimensional local fit and compresses only cross-party coupling. Let $\mathbf{A}_k=\mathbf{X}_k^\top \mathbf{X}_k+\gamma \mathbf{I}$, $\mathbf{H}_k=\mathbf{X}_k^\top \mathbf{Y}$, and $\mathbf{M}_k$ be a local spectral map normalized by $\mathbf{M}_k^\top \mathbf{A}_k\mathbf{M}_k=\mathbf{I}$. After a reduced coupled solve, party $k$ recovers
\begin{equation}
\mathbf{W}_k^{\mathrm{LC}}=\mathbf{A}_k^{-1}\mathbf{H}_k+\mathbf{M}_k(\mathbf{u}_k-\mathbf{b}_k),\qquad \mathbf{b}_k=\mathbf{M}_k^\top \mathbf{H}_k,
\label{eq:lc-recovery}
\end{equation}
where $\mathbf{u}_k$ is its block of the coupled solution. Subtracting $\mathbf{b}_k$ avoids counting the retained local component twice. LC retains full-width local work and is approximate at reduced ranks. Full-rank maps recover the original solution.

These computations reuse the protected matrix interfaces above under the same security assumptions. Section~\ref{subsec:Validity Analysis} states their exactness and approximation properties, with proofs in Appendix~C.5. Appendix~G gives their implementation steps.

\subsection{Theoretical Analyses}
\label{Sec:Theoretical-Analyses}

In this section, we present comprehensive theoretical analyses of the proposed VAFL, encompassing validity analyses, privacy analyses, and efficiency analyses.
Specifically, we theoretically demonstrate that the local models derived via VAFL are mathematically equivalent to the optimal solution obtained through centralized training on the complete dataset. 
These equivalence claims concern full-dimensional VAFL and are retained by VAFL-E. The first analysis part also establishes the projected optimum, LC surrogate characterization and approximation-error bound for the reduced-rank variants. We also characterize ciphertext-component confidentiality, private-coordinate identities and the operation costs of the implemented workflow. 
Detailed privacy analyses are provided in Appendix~D.

\subsubsection{Validity Analyses}
\label{subsec:Validity Analysis}
Here, we present comprehensive validity analyses of our proposed VAFL, theoretically demonstrating that the local models derived via our VAFL are mathematically equivalent to the optimal solution achieved through centralized training.
To this end, in Lemma~1, we first establish the closed-form expression for the optimal model achievable through centralized training.
Subsequently, in Theorem~1 and Theorem~2, we prove that the results obtained from our VAFL centralized workflow for the two-party and multi-party cases, respectively, are identical to the optimal models.
Theorems~3 and~4 then establish coefficient recovery under orthogonal feature transformations and invertible target transformations, respectively, completing the algebraic validity argument for the federated implementation. We then establish the validity and approximation properties of the acceleration strategies in Propositions~\ref{prop:accel-exact-projection}--\ref{prop:accel-lc-error}.

First, let's focus on deriving the optimal solution for the optimization objective~\eqref{eq:Section-3-1-1}, which also fundamentally represents the optimal complete model achievable through centralized training on the complete dataset.
The detailed derivation is established in Lemma 1 as follows:

\noindent\textbf{Lemma 1:} Consider the optimization problem:
\begin{equation}
\label{eq:Lemma-1-1}
    \mathop{\arg\min}\limits_{\mathbf{W}} {\left\|\mathbf{Y} -\mathbf{X} \mathbf{W} \right\|}_F^2+\gamma{\left\| \mathbf{W} \right\|}_F^2,
\end{equation}
and the closed-form solution of this problem is:
\begin{equation}
\label{eq:Lemma-1-2}
    \mathbf{\hat W} = {({\mathbf{X}}^\top\mathbf{X}+\gamma\mathbf{I})}^{-1}{\mathbf{X}}^\top\mathbf{Y}.
\end{equation}

\noindent\textit{\textbf{Proof.}} 
This result follows from the standard derivation of the least-squares method with $\ell_2$ regularization. For the detailed derivation, please refer to Appendix~C.1.

Second, we further demonstrate that the local models derived from the centralized workflow of our VAFL are identical to the optimal models obtained via centralized training on the complete dataset, whose closed-form solution is given by Eq.~\eqref{eq:Lemma-1-2} in Lemma~1.
Specifically, we begin by analyzing the fundamental two-party case, and the detailed analysis is presented in Theorem 1 as follows:

\noindent\textbf{Theorem 1:} Consider a VFL system with two parties, where the complete feature matrix is vertically partitioned as $\mathbf{X} = [\mathbf{X}_1, \mathbf{X}_2] $ corresponding to the two parties, and the label matrix is $\mathbf{Y}$.
The closed-form expressions for the sub-components of the optimal complete model $\mathbf{\hat W} = [\mathbf{\hat W}_1^\top, \mathbf{\hat W}_2^\top]^\top$, which minimizes the empirical risk objective \eqref{eq:Section-3-1-1} on the complete dataset $\{ \mathbf{X}, \mathbf{Y} \}$, are given by:
\begin{equation}
\label{eq:Theorem-1-1}
    \begin{cases}
        \mathbf{\hat W}_1 = \mathbf{U}_1\mathbf{L}_1 - (\mathbf{F}_{1,1} + \gamma\mathbf{I})^{-1} \mathbf{F}_{1,2} \mathbf{U}_2\mathbf{L}_2, \\
        \mathbf{\hat W}_2 = \mathbf{U}_2\mathbf{L}_2 - (\mathbf{F}_{2,2} + \gamma\mathbf{I})^{-1} \mathbf{F}_{2,1} \mathbf{U}_1\mathbf{L}_1,
    \end{cases}
\end{equation}
where
\begin{equation}
\label{eq:Theorem-1-2}
    \begin{cases}
        \mathbf{U}_1 = [(\mathbf{F}_{1,1} + \gamma\mathbf{I}) - \mathbf{F}_{1,2} (\mathbf{F}_{2,2} + \gamma\mathbf{I})^{-1} \mathbf{F}_{2,1}]^{-1},\\
        \mathbf{U}_2 = [(\mathbf{F}_{2,2} + \gamma\mathbf{I}) - \mathbf{F}_{2,1} (\mathbf{F}_{1,1} + \gamma\mathbf{I})^{-1} \mathbf{F}_{1,2}]^{-1},  \\
    \end{cases}
    \quad
    \begin{cases}
         \mathbf{F}_{i,j} = \mathbf{X}_i^\top\mathbf{X}_j,\\
        \mathbf{L}_i=\mathbf{X}_i^\top\mathbf{Y}. \\
    \end{cases}
\end{equation}

\noindent\textit{\textbf{Proof.}}
As established in Lemma 1, the closed-form expression for the optimal complete model $\mathbf{\hat W}$ is given by \eqref{eq:Lemma-1-2}. 
By substituting partitioned form $\mathbf{X} = [\mathbf{X}_1, \mathbf{X}_2]$ into \eqref{eq:Lemma-1-2} and applying the \textit{Matrix Block Inversion} Lemma, we derive the explicit formulas presented in \eqref{eq:Theorem-1-1} and \eqref{eq:Theorem-1-2}. 
For the detailed derivation, please refer to Appendix~C.2.

Meanwhile, we analyze the multi-party case in Theorem 2, proving that the local models derived via \eqref{eq:section-4-1-9}--\eqref{eq:section-4-1-11} are mathematically equivalent to the optimal solution. 
The detailed proof is presented as follows:

\noindent\textbf{Theorem 2:}
Consider a VFL system with $K$ parties, where the complete feature matrix is vertically partitioned as $\mathbf{X} = [\mathbf{X}_1, \mathbf{X}_2, \cdots, \mathbf{X}_K] $ corresponding to the $K$ parties, and the label matrix is $\mathbf{Y}$.
The corresponding optimal complete model $\mathbf{\hat W}$, which minimizes the empirical risk objective \eqref{eq:Section-3-1-1} on the complete dataset $\{ \mathbf{X}, \mathbf{Y} \}$, is given by \eqref{eq:Theorem-2-1}. 
Furthermore, the optimal local models constitute a sub-component of this complete model and can be obtained by partitioning the complete model $\mathbf{\hat W} = [\mathbf{\hat W}_{1}^\top, \mathbf{\hat W}_{2}^\top, \cdots, \mathbf{\hat W}_{K}^\top]^\top$ accordingly.
\begin{equation}
\label{eq:Theorem-2-1}
    \mathbf{\hat W} = \mathbf{R}_K [\mathbf{L}_{1}^\top, \mathbf{L}_{2}^\top, \cdots, \mathbf{L}_{K}^\top]^\top,
\end{equation}
where the \textit{Aggregated Gram Inverse} $\mathbf{R}_K$ is derived by recursively executing the following formulations:
\begin{equation}
\mathbf{R}_k = 
\begin{bmatrix}
    \mathbf{R}_k^{(1)} & \mathbf{R}_k^{(2)} \\
    \mathbf{R}_k^{(3)} & \mathbf{R}_k^{(4)}
\end{bmatrix},
\qquad
\label{eq:Theorem-2-2}
\begin{cases}
\mathbf{R}^{(1)}_k = \mathbf{R}_{k-1} + \mathbf{R}_{k-1} \mathbf{F}_{1:k-1,k} \mathbf{R}^{(4)}_k \mathbf{F}_{k,1:k-1} \mathbf{R}_{k-1}, \\
\mathbf{R}^{(2)}_k = -\mathbf{R}_{k-1}\mathbf{F}_{1:k-1,k}\mathbf{R}^{(4)}_k = \mathbf{R}^{(3)\top}_k,\\
\mathbf{R}^{(3)}_k = -(\mathbf{F}_{k,k} + \gamma \mathbf{I})^{-1} \mathbf{F}_{k,1:k-1} \mathbf{R}^{(1)}_k = \mathbf{R}^{(2)\top}_k,\\
\mathbf{R}^{(4)}_k = ( \mathbf{F}_{k,k} + \gamma \mathbf{I} - \mathbf{F}_{k,1:k-1} \mathbf{R}_{k-1} \mathbf{F}_{1:k-1,k} )^{-1}.
\end{cases}
\end{equation}

\noindent\textit{\textbf{Proof.}}
The proof involves two primary steps: (1) Leveraging mathematical induction and the Woodbury matrix identity, we derive the closed-form expression for the recursive term $\mathbf{R}_k$. 
(2) By substituting this derived expression into \eqref{eq:Theorem-2-1}, we demonstrate that the resulting complete model $\mathbf{\hat W}$ mathematically satisfies the optimality condition established in \eqref{eq:Lemma-1-2}. 
For the detailed derivation, please refer to Appendix~C.3.

Third, considering the federated implementation of our VAFL, wherein each party $k$ introduces a private random orthogonal matrix $\bm{\Pi}_k$ to mask sensitive information, we demonstrate in Theorem~3 that each party can remove the feature-coordinate transformation by left-multiplying its transformed local coefficient block by $\bm{\Pi}_k$. Theorem~4 establishes the corresponding recovery for an invertible target transformation.
Consequently, by synthesizing this result with Theorem 1 and Theorem 2, we can prove that the local models derived via VAFL in the VFL setting are equivalent to the optimal models. 
The detailed proof is presented as follows:

\noindent\textbf{Theorem 3:}
For any complete dataset $\{\mathbf{X} \in \mathbb{R}^{\text N \times \widetilde{\text F}},\mathbf{Y} \in \mathbb{R}^{\text N \times {\text C}}\}$ and any orthogonal matrix $\bm{\Pi} \in \mathbb{R}^{\widetilde{\text F} \times \widetilde{\text F}}$, let the perturbed dataset be $\{\mathbf{X}', \mathbf{Y}\}$ where $\mathbf{X}' = \mathbf{X}\bm{\Pi}$.
Similarly, let $\mathbf{W}$ and $\mathbf{W}'$ denote the closed-form solutions obtained from the original and perturbed datasets according to Eq.~\eqref{eq:Lemma-1-2}, respectively.
Then, the following relationship holds:
\begin{equation}
\label{eq:Lemma-4-1}
\mathbf W = \bm{\Pi}\mathbf W'.
\end{equation}

\noindent\textit{\textbf{Proof.}}
The validity of this relationship is established by leveraging the fundamental properties of orthogonal matrices (specifically $\bm{\Pi}^\top = \bm{\Pi}^{-1}$). 
For the detailed derivation, please refer to Appendix~C.4.

\noindent\textbf{Theorem 4:}
For any complete dataset $\{\mathbf{X} \in \mathbb{R}^{\text N \times \widetilde{\text F}},\mathbf{Y} \in \mathbb{R}^{\text N \times {\text C}}\}$ and any invertible matrix $\bm{\Pi} \in \mathbb{R}^{\text C \times \text C}$, let the perturbed dataset be $\{\mathbf{X}, \mathbf{Y}'\}$ where $\mathbf{Y}' = \mathbf{Y}\bm{\Pi}$.
Similarly, let $\mathbf{W}$ and $\mathbf{W}'$ denote the closed-form solutions obtained from the original and perturbed datasets according to Eq.~\eqref{eq:Lemma-1-2}, respectively.
Then, the following relationship holds:
\begin{equation}
\label{eq:Theorem-4-label-recovery}
\mathbf W = \mathbf W'\bm{\Pi}^{-1}.
\end{equation}

\noindent\textit{\textbf{Proof.}}
The validity of this relationship is established by leveraging the fundamental properties of invertible matrices (specifically $\bm{\Pi}\bm{\Pi}^{-1} = \mathbf{I}$). 
For the detailed derivation, please refer to Appendix~C.4.

\paragraph{\textbf{\upshape Validity and approximation of the acceleration strategies}}
We now extend the validity analysis to the strategies in Section~\ref{sec:acceleration}. For fixed supplied representations $\mathbf{X}=[\mathbf{X}_1,\ldots,\mathbf{X}_K]$ and $\gamma>0$, write $\mathbf{G}=\mathbf{X}^\top \mathbf{X}+\gamma \mathbf{I}$, $\mathbf{H}=\mathbf{X}^\top \mathbf{Y}$, $\mathbf{A}_k=\mathbf{X}_k^\top \mathbf{X}_k+\gamma \mathbf{I}$, and $\mathbf{W}^\star=\mathbf{G}^{-1}\mathbf{H}$. The following results concern exact arithmetic and the same representations, targets and regularization.

\renewcommand{\thetheorem}{\arabic{theorem}}
\refstepcounter{theorem}\label{prop:accel-exact-projection}
\noindent\textbf{Proposition~\thetheorem\ (equivalent solving and projection):}
VAFL-E recovers $\mathbf{W}^\star$ by applying the block factors of $\mathbf{G}$ to $\mathbf{H}$. For any block-diagonal orthonormal map $\mathbf{Q}$ with $\mathbf{Q}^\top \mathbf{Q}=\mathbf{I}$, Eq.~\eqref{eq:sp-lift} gives the unique Ridge optimum subject to $\mathbf{W}\in\operatorname{range}(\mathbf{Q})$ and preserves its predictions under lifting: $\mathbf{X}\mathbf{W}^{\mathrm{proj}}=\mathbf{X}\mathbf{Q}\mathbf{U}=\mathbf{Z}\mathbf{U}$. If $\mathbf{Q}$ is square, this optimum equals $\mathbf{W}^\star$.

\noindent\textit{\textbf{Proof.}}
Block substitution yields $\mathbf{G}\mathbf{W}=\mathbf{H}$, and orthonormal projection preserves the Ridge penalty within the retained subspace. The complete arguments are in Appendices~C.5.1 and~C.5.2.

\refstepcounter{theorem}\label{prop:accel-lc-surrogate}
\noindent\textbf{Proposition~\thetheorem\ (LC surrogate and full-rank recovery):}
Let $\mathbf{A}=\operatorname{diag}(\mathbf{A}_1,\ldots,\mathbf{A}_K)$ and $\mathbf{M}=\operatorname{diag}(\mathbf{M}_1,\ldots,\mathbf{M}_K)$, with full-column-rank local maps satisfying $\mathbf{M}_k^\top \mathbf{A}_k\mathbf{M}_k=\mathbf{I}$. Then $\mathbf{C}=\mathbf{M}^\top \mathbf{G}\mathbf{M}$ is positive definite, and the LC model in Eq.~\eqref{eq:lc-recovery} is the unique solution of $\widehat{\mathbf{G}} \mathbf{W}=\mathbf{H}$, where $\widehat{\mathbf{G}}\succ0$ has blocks
\begin{equation}
\widehat{\mathbf{G}}_{kk}=\mathbf{A}_k,\qquad
\widehat{\mathbf{G}}_{ij}=\mathbf{A}_i\mathbf{M}_i\mathbf{C}_{ij}\mathbf{M}_j^\top \mathbf{A}_j\quad(i\ne j).
\label{eq:lc-surrogate}
\end{equation}
If every local map is square, $\widehat{\mathbf{G}}=\mathbf{G}$ and LC recovers $\mathbf{W}^\star$.

\noindent\textit{\textbf{Proof.}}
An orthogonal decomposition in the locally normalized coordinates gives the surrogate inverse and the coefficient-recovery formula. Appendix~C.5.3 supplies the full derivation.

\refstepcounter{theorem}\label{prop:accel-lc-error}
\noindent\textbf{Proposition~\thetheorem\ (LC approximation error):}
Under the conditions of Proposition~\ref{prop:accel-lc-surrogate}, the reduced-rank LC model satisfies
\begin{equation}
\|\mathbf{W}^{\mathrm{LC}}-\mathbf{W}^\star\|_F
\le\frac{\|\mathbf{G}-\widehat{\mathbf{G}}\|_2}{\lambda_{\min}(\widehat{\mathbf{G}})}\,\|\mathbf{W}^\star\|_F.
\label{eq:lc-error-bound}
\end{equation}
The corresponding prediction discrepancy $\|\mathbf{X}(\mathbf{W}^{\mathrm{LC}}-\mathbf{W}^\star)\|_F$ is bounded by the same right-hand side multiplied by $\|\mathbf{X}\|_2$.

\noindent\textit{\textbf{Proof.}}
Subtract the full and surrogate normal equations, then apply the spectral norm bound for $\widehat{\mathbf{G}}^{-1}$. Appendix~C.5.4 gives the error identity and norm derivation.

The error bound depends on both the omitted coupling $\mathbf{G}-\widehat{\mathbf{G}}$ and the conditioning of $\widehat{\mathbf{G}}$. It characterizes coefficient and score discrepancies for fixed representations. Retained variance or a small solve residual alone does not guarantee preserved predictive accuracy.

\subsubsection{Privacy Analyses}
\label{subsec:Privacy Analyses}

\renewcommand{\thetheorem}{\arabic{theorem}}

Our analysis focuses on the protection of local feature information and raw input records specified in the privacy objective. VAFL keeps these matrices at their owners and uses encrypted computations and private local coordinate transformations for collaboration. We establish two supporting properties: confidentiality of ciphertext components from observers without the secret key, and algebraic ambiguity in the original feature coordinates.

\refstepcounter{theorem}\label{prop:privacy-messages}
\noindent\textbf{Proposition~\thetheorem:} Assume that Paillier is semantically secure. For plaintext matrices with the same public dimensions and encoding metadata, their fresh encryptions are computationally indistinguishable to a polynomial-time observer without the secret key. This ciphertext-component property also applies to coalitions without the key.

\noindent\textit{\textbf{Proof.}}
Replacing an encrypted component by a fresh encryption of a dummy plaintext with the same public layout gives the standard semantic-security experiment. Appendix~D.1 distinguishes this component argument from a simulation of the full view, including plaintext messages and assigned outputs.

\refstepcounter{theorem}\label{prop:privacy-rotation}
\noindent\textbf{Proposition~\thetheorem:} Let $\mathbf Z_k=\mathbf{\widetilde X}_k\bm\Pi_k$, where $\bm\Pi_k$ is orthogonal. For any orthogonal $\mathbf Q$ of matching dimension,
\begin{equation}
\label{eq:privacy-coordinate-ambiguity}
\begin{aligned}
\mathbf Z_k&=(\mathbf{\widetilde X}_k\mathbf Q)(\mathbf Q^\top\bm\Pi_k),\\
\mathbf Z_k\mathbf Z_k^\top&=\mathbf{\widetilde X}_k\mathbf{\widetilde X}_k^\top.
\end{aligned}
\end{equation}
These identities describe alternative coordinate factorizations and the preserved row Gram matrix.

\noindent\textit{\textbf{Proof.}}
Both identities follow from orthogonality. Appendix~D.2 explains the resulting coordinate ambiguity, the preserved statistics and the role of additional information.

Together, these properties support the feature and raw-input protection objective without directly pooling either matrix. Appendix~D.1.1 analyzes the remaining algebraic ambiguity even when an observer obtains a Gram matrix. Appendix~D states the conditions of this reconstruction analysis and accounts for the full set of relevant plaintext observations and available auxiliary information.

\subsubsection{Efficiency Analyses}
\label{subsec:Efficiency Analyses}

We summarize the computational and communication costs of VAFL below. Let $D=\sum_{k=1}^{K}\widetilde{\text F}_k$ be the total supplied feature width, $\text C$ the output width, and $d=D+\text C$. Plaintext scalar work is normalized to one. The costs of encryption, decryption, homomorphic addition and homomorphic scalar multiplication are denoted by $T_{\text{Enc}}$, $T_{\text{Dec}}$, $T_{\text{Add}}$ and $T_{\text{Mul}}$, respectively, and $\text S$ is the ciphertext size in plaintext units. Full protocol-level counts and the two-party and multi-party derivations are deferred to Appendix~E.

\refstepcounter{theorem}\label{prop:efficiency-complexity}
\noindent\textbf{Proposition~\thetheorem:} For the dense-matrix operation model of the stated VAFL workflow, the aggregate work across the parties and agent admits the upper bounds
\begin{equation}
\label{eq:efficiency-aggregate-bounds}
    \begin{cases}
        \mathcal{C}_{\text{VAFL}} = \mathcal{O}(K\text N d^2 + Kd^3(1+T_{\text{Mul}}+T_{\text{Add}}) + Kd^2(T_{\text{Enc}}+T_{\text{Dec}})),\\
        \mathcal{V}_{\text{VAFL}} = \mathcal{O}(K\text N d + Kd^2\text S).
    \end{cases}
\end{equation}
Here, $\mathcal{C}_{\text{VAFL}}$ counts computational work and $\mathcal{V}_{\text{VAFL}}$ counts communicated plaintext/ciphertext units.

\noindent\textit{\textbf{Proof.}}
Summing the operation and message counts for Protocols 1--3 and the five training stages gives the bounds. The aggregation and sharper participant-specific expressions appear in Appendix~E.1 and Appendix Tables~4 and~5.

\refstepcounter{theorem}\label{prop:efficiency-sample-scaling}
\noindent\textbf{Proposition~\thetheorem:} For fixed $K$, $D$ and $\text C$, VAFL's feature-space inversion and model-solution dimensions do not depend on the number of records $\text N$. Sample-dependent work enters through feature and target product formation, while the analytic solve requires no training-epoch factor. Replacing the supplied width $D$ by a projected width $m<D$ reduces the feature-dependent quadratic and cubic terms of the subsequent solve.

\noindent\textit{\textbf{Proof.}}
The sufficient products have dimensions $D\times D$ and $D\times\text C$, and the block recursion operates on their feature partitions. The detailed dimension and operation-count argument is given in Appendix~E.2.

This separation explains why fixed-width analytic training can incorporate additional records without enlarging its matrix inverse, and why local projection provides an acceleration option for wide tabular heads. Fitting efficiency and sample-size scaling are examined in Sections~\ref{sec:exp-efficiency} and~\ref{sec:exp-training-size}. The symbolic bounds count operations and communication units.

\paragraph{\textbf{\upshape Costs of accelerated solving}}
For total width $D$ and output width $c$, VAFL-E retains dense $\mathcal O(D^3)$ factorization and $\mathcal O(D^2)$ factor storage, followed by $\mathcal O(D^2c)$ target solves. RP/SP reduce the coupled width to $r<D$. LC also reduces the coupled factorization to width $r$, while retaining full-width local factorizations and coefficient applications. Appendix~E.4 details these operation costs.

\section{Experimental Evaluation}
\label{Experiments}

We compare predictive effectiveness on five benchmarks, then examine training-data scaling, acceleration strategies, party scalability, regularization sensitivity and frozen-backbone compatibility. The main table reports accuracy and RMSE. Fitting cost is evaluated in the dedicated training-data scaling, acceleration and party-scaling experiments.

\subsection{Experimental Setup}
\label{sec:exp-setup}
\paragraph{\textbf{\upshape Datasets}}
We use three classification datasets, CIFAR-10, CIFAR-100 and Multiple Features (MF), and two regression datasets, Diabetes and Ailerons. The two CIFAR datasets each contain 60,000 images and preserve the official 50,000/10,000 development/test division. We reserve 5,000 development images for validation, leaving 45,000 head-training images. CIFAR-10 has ten classes, while CIFAR-100 has 100 fine-grained classes~\cite{exp-cifar}. MF contains 2,000 handwritten digits with 649 supplied numerical predictors formed by concatenating six aligned feature groups~\cite{choice-mfeat}. Its training/validation/test counts are 1,440/160/400. These datasets compare the methods on both learned image features and high-dimensional numerical features.

The regression tasks vary substantially in sample count and feature dimension. Diabetes has 442 records and ten clinical predictors, with 317/36/89 training/validation/test records~\cite{exp-sklearn}. Ailerons has 13,750 records and 40 raw state variables, with a 9,900/1,100/2,750 split~\cite{choice-ailerons}. We retain all labeled records in these benchmarks. All compared methods share this split policy. Appendix~F.1 gives the feature descriptions and complete split policy.

\paragraph{\textbf{\upshape Baselines}}
We compare VAFL with four baselines in the same vertical federated learning framework. \textbf{Linear} is an additive linear predictor that sums the parties' local logits. \textbf{XGB} is a vertical histogram gradient-boosting method using regularized second-order split statistics~\cite{exp-xgboost}. \textbf{RF} is a vertical histogram random forest that averages predictions from bootstrap-trained trees~\cite{exp-random-forest}. \textbf{NN} is a split multilayer perceptron with local bottom networks and an active-party top network. All four baselines share VAFL's sample IDs, data splits and original supplied features within each setting. The timing comparisons in Sections~\ref{sec:exp-training-size} and~\ref{sec:exp-acceleration} use the corresponding CPU-based Paillier-assisted protocols. Appendix~F.2 gives their full definitions and implementation details.

\paragraph{\textbf{\upshape VAFL variants and metrics}}
We compare original full-dimensional VAFL, its equivalent solver VAFL-E, and VAFL$_m$ at total non-bias budgets $m\in\{256,128,64\}$. VAFL-E preserves the original Ridge optimum in exact arithmetic. Unless a strategy is explicitly named, VAFL$_m$ uses LC on CIFAR-10/100 and SP on the tabular datasets, including the training-data scaling experiment. RP is a matched-width reference in the acceleration analysis. Diabetes retains its ten native fields and is an identity control. LC uses validation-selected regularization on CIFAR. Original VAFL and SP retain their specified fixed penalties. Appendix~F.3 specifies the representations and width allocation, and Appendix~H gives the complete compression results.

Predictive performance is measured in plaintext. We report mean $\pm$ SEM in tables and mean curves without error bars in figures. Classification uses accuracy (\%). Regression uses RMSE, with Ailerons multiplied by $10^4$. Bold marks the highest mean accuracy or lowest mean RMSE in each main-table row. The shared VAFL (VAFL-E) column reports the original full-dimensional predictor reference; VAFL-E has the same optimum in exact arithmetic. The training-data experiment also checks their numerical agreement directly. Training time is analyzed separately in the scaling and acceleration experiments.

\subsection{RQ1: Predictive Effectiveness}
\label{sec:exp-effectiveness}
\begin{table}[tb]
\begingroup\centering
\captionsetup{font={small,rm},labelfont={rm},textfont={rm}}
\caption{Predictive effectiveness on five benchmarks. Classification reports accuracy (\%), and regression reports RMSE (Ailerons scaled by $10^4$). Values are mean $\pm$ SEM; bold marks the best mean in each row. VAFL$_m$ uses LC on CIFAR and SP on tabular data.}
\label{tab:exp-main}
\normalfont\rmfamily\scriptsize
\renewcommand{\arraystretch}{1.20}
\setlength{\tabcolsep}{3pt}
\setlength{\arrayrulewidth}{0.4pt}
\resizebox{\textwidth}{!}{\begin{tabular}{c|c|c|*{8}{c}}
\hline
\multirow{2}{*}{Task} & \multirow{2}{*}{Dataset} & \multirow{2}{*}{$K$} & \multicolumn{8}{c}{Accuracy (\%) $\uparrow$} \\
\cline{4-11}
 &  &  & Linear & XGB & RF & NN & VAFL (VAFL-E) & VAFL$_{256}$ & VAFL$_{128}$ & VAFL$_{64}$ \\
\hline
\multirow[c]{12}{*}{CLS} & \multirow[c]{4}{*}{CIFAR-10} & $2$ & $67.44_{\pm 0.44}$ & $59.97_{\pm 0.11}$ & $47.33_{\pm 0.02}$ & $65.99_{\pm 0.13}$ & $\mathbf{68.23_{\pm 0.01}}$ & $68.06_{\pm 0.08}$ & $67.55_{\pm 0.04}$ & $67.26_{\pm 0.05}$ \\
 &  & $4$ & $72.27_{\pm 0.24}$ & $64.45_{\pm 0.11}$ & $51.27_{\pm 0.10}$ & $74.14_{\pm 0.34}$ & $\mathbf{75.48_{\pm 0.08}}$ & $74.07_{\pm 0.09}$ & $73.23_{\pm 0.05}$ & $72.99_{\pm 0.02}$ \\
 &  & $8$ & $59.71_{\pm 0.34}$ & $55.23_{\pm 0.15}$ & $41.71_{\pm 0.12}$ & $62.85_{\pm 0.19}$ & $\mathbf{65.16_{\pm 0.05}}$ & $63.22_{\pm 0.02}$ & $62.63_{\pm 0.06}$ & $61.91_{\pm 0.05}$ \\
 &  & $16$ & $66.82_{\pm 0.10}$ & $60.63_{\pm 0.09}$ & $45.19_{\pm 0.14}$ & $68.62_{\pm 0.08}$ & $\mathbf{70.55_{\pm 0.06}}$ & $66.45_{\pm 0.04}$ & $65.71_{\pm 0.11}$ & $64.50_{\pm 0.10}$ \\
\cline{2-11}
 & \multirow[c]{4}{*}{CIFAR-100} & $2$ & $38.42_{\pm 0.07}$ & $30.55_{\pm 0.15}$ & $16.18_{\pm 0.24}$ & $32.50_{\pm 0.12}$ & $\mathbf{41.49_{\pm 0.08}}$ & $41.23_{\pm 0.04}$ & $41.11_{\pm 0.06}$ & $40.92_{\pm 0.09}$ \\
 &  & $4$ & $46.17_{\pm 0.28}$ & $33.81_{\pm 0.22}$ & $14.78_{\pm 0.34}$ & $37.72_{\pm 0.16}$ & $\mathbf{49.12_{\pm 0.04}}$ & $47.51_{\pm 0.03}$ & $47.00_{\pm 0.05}$ & $46.49_{\pm 0.06}$ \\
 &  & $8$ & $32.46_{\pm 0.06}$ & $26.82_{\pm 0.23}$ & $14.12_{\pm 0.25}$ & $25.92_{\pm 0.37}$ & $\mathbf{36.64_{\pm 0.07}}$ & $34.28_{\pm 0.16}$ & $33.68_{\pm 0.11}$ & $32.87_{\pm 0.05}$ \\
 &  & $16$ & $39.98_{\pm 0.06}$ & $30.69_{\pm 0.26}$ & $15.14_{\pm 0.04}$ & $30.81_{\pm 0.16}$ & $\mathbf{42.40_{\pm 0.08}}$ & $38.01_{\pm 0.06}$ & $36.83_{\pm 0.08}$ & $35.70_{\pm 0.06}$ \\
\cline{2-11}
 & \multirow[c]{4}{*}{MF} & $2$ & $99.00_{\pm 0.00}$ & $98.50_{\pm 0.25}$ & $97.33_{\pm 0.33}$ & $98.25_{\pm 0.25}$ & $\mathbf{99.08_{\pm 0.08}}$ & $98.75_{\pm 0.14}$ & $98.75_{\pm 0.14}$ & $98.75_{\pm 0.14}$ \\
 &  & $4$ & $98.92_{\pm 0.08}$ & $98.58_{\pm 0.22}$ & $97.08_{\pm 0.30}$ & $99.00_{\pm 0.14}$ & $\mathbf{99.08_{\pm 0.08}}$ & $98.83_{\pm 0.30}$ & $98.75_{\pm 0.00}$ & $97.92_{\pm 0.30}$ \\
 &  & $8$ & $98.83_{\pm 0.22}$ & $98.17_{\pm 0.08}$ & $96.92_{\pm 0.22}$ & $99.00_{\pm 0.25}$ & $\mathbf{99.08_{\pm 0.08}}$ & $98.58_{\pm 0.08}$ & $98.75_{\pm 0.14}$ & $97.75_{\pm 0.14}$ \\
 &  & $16$ & $98.83_{\pm 0.08}$ & $98.83_{\pm 0.08}$ & $97.17_{\pm 0.22}$ & $99.00_{\pm 0.00}$ & $99.08_{\pm 0.08}$ & $\mathbf{99.25_{\pm 0.14}}$ & $98.58_{\pm 0.08}$ & $97.92_{\pm 0.17}$ \\
\hline
\multirow{2}{*}{Task} & \multirow{2}{*}{Dataset} & \multirow{2}{*}{$K$} & \multicolumn{8}{c}{RMSE $\downarrow$} \\
\cline{4-11}
 &  &  & Linear & XGB & RF & NN & VAFL (VAFL-E) & VAFL$_{256}$ & VAFL$_{128}$ & VAFL$_{64}$ \\
\hline
\multirow[c]{7}{*}{REG} & \multirow[c]{3}{*}{Diabetes} & $2$ & $61.938_{\pm 3.589}$ & $64.037_{\pm 1.065}$ & $59.901_{\pm 1.252}$ & $60.169_{\pm 1.524}$ & $\mathbf{56.442_{\pm 1.176}}$ & $\mathbf{56.442_{\pm 1.176}}$ & $\mathbf{56.442_{\pm 1.176}}$ & $\mathbf{56.442_{\pm 1.176}}$ \\
 &  & $4$ & $64.345_{\pm 4.735}$ & $65.721_{\pm 1.880}$ & $61.165_{\pm 1.958}$ & $60.219_{\pm 0.277}$ & $\mathbf{56.442_{\pm 1.176}}$ & $\mathbf{56.442_{\pm 1.176}}$ & $\mathbf{56.442_{\pm 1.176}}$ & $\mathbf{56.442_{\pm 1.176}}$ \\
 &  & $8$ & $80.842_{\pm 8.304}$ & $65.721_{\pm 1.880}$ & $61.167_{\pm 1.866}$ & $60.533_{\pm 0.110}$ & $\mathbf{56.442_{\pm 1.176}}$ & $\mathbf{56.442_{\pm 1.176}}$ & $\mathbf{56.442_{\pm 1.176}}$ & $\mathbf{56.442_{\pm 1.176}}$ \\
\cline{2-11}
 & \multirow[c]{4}{*}{Ailerons} & $2$ & $1.627_{\pm 0.019}$ & $1.601_{\pm 0.004}$ & $1.702_{\pm 0.008}$ & $1.662_{\pm 0.028}$ & $\mathbf{1.581_{\pm 0.004}}$ & $1.583_{\pm 0.002}$ & $1.585_{\pm 0.001}$ & $1.621_{\pm 0.005}$ \\
 &  & $4$ & $1.612_{\pm 0.004}$ & $1.601_{\pm 0.007}$ & $1.706_{\pm 0.010}$ & $1.683_{\pm 0.027}$ & $\mathbf{1.581_{\pm 0.004}}$ & $1.582_{\pm 0.001}$ & $1.595_{\pm 0.003}$ & $1.637_{\pm 0.009}$ \\
 &  & $8$ & $1.610_{\pm 0.003}$ & $1.601_{\pm 0.007}$ & $1.704_{\pm 0.010}$ & $1.680_{\pm 0.011}$ & $\mathbf{1.581_{\pm 0.004}}$ & $1.584_{\pm 0.002}$ & $1.604_{\pm 0.001}$ & $1.652_{\pm 0.010}$ \\
 &  & $16$ & $1.619_{\pm 0.006}$ & $1.611_{\pm 0.001}$ & $1.702_{\pm 0.009}$ & $1.678_{\pm 0.020}$ & $\mathbf{1.581_{\pm 0.004}}$ & $1.586_{\pm 0.002}$ & $1.618_{\pm 0.005}$ & $1.764_{\pm 0.023}$ \\
\hline
\end{tabular}}
\endgroup
\end{table}

Table~\ref{tab:exp-main} compares predictive effectiveness across the five benchmarks. Full-dimensional VAFL has the highest observed accuracy or lowest RMSE among the five full-input methods in all 19 dataset--party settings. On CIFAR-10 and CIFAR-100 at $K=4$, its accuracies are 75.48\% and 49.12\%, compared with the strongest baseline results of 74.14\% and 46.17\%. MF reaches 99.08\% across party counts. Diabetes and Ailerons retain RMSEs of 56.442 and 1.581. These observations compare the specified predictors and training settings, rather than establishing an advantage over an equivalent direct Ridge optimum.

VAFL-E reorganizes the full-dimensional solve without changing its optimum in exact arithmetic, while the reduced variants also change the fitted prediction head. At $K=2,m=256$, CIFAR LC reaches 68.06\% and 41.23\%, within 0.17 and 0.26 percentage points of the full-dimensional references. On MF and Ailerons, SP$_{256}$ gives 98.75\% and RMSE 1.583, respectively. Compression losses depend on the dataset, budget and party count. For example, at $K=16$, CIFAR-100 accuracy falls from 42.40\% for full-dimensional VAFL to 38.01\% at $m=256$ and 35.70\% at $m=64$. Diabetes remains unchanged because all ten native features are retained. Fitting cost is examined separately in Sections~\ref{sec:exp-training-size} and~\ref{sec:exp-acceleration}.

\subsection{RQ2: Impact of Training Data Size}
\label{sec:exp-training-size}
We vary the training fraction over 25\%, 50\%, 75\% and 100\% on CIFAR-10, CIFAR-100 and Ailerons at $K=2$. All methods use the same nested training subsets, with fixed validation and test partitions. CIFAR standardization and the local spectral bases are fitted on the current training subset only. The supplied feature extractors remain frozen. Baseline architectures and training budgets are unchanged. VAFL$_{256}$ uses LC on CIFAR and SP on Ailerons. Each CIFAR fraction follows the same validation-only regularization selection rule. Appendix~F.2 provides the sampling and numerical checks.

\begin{figure}[H]
\centering
\includegraphics[width=\linewidth]{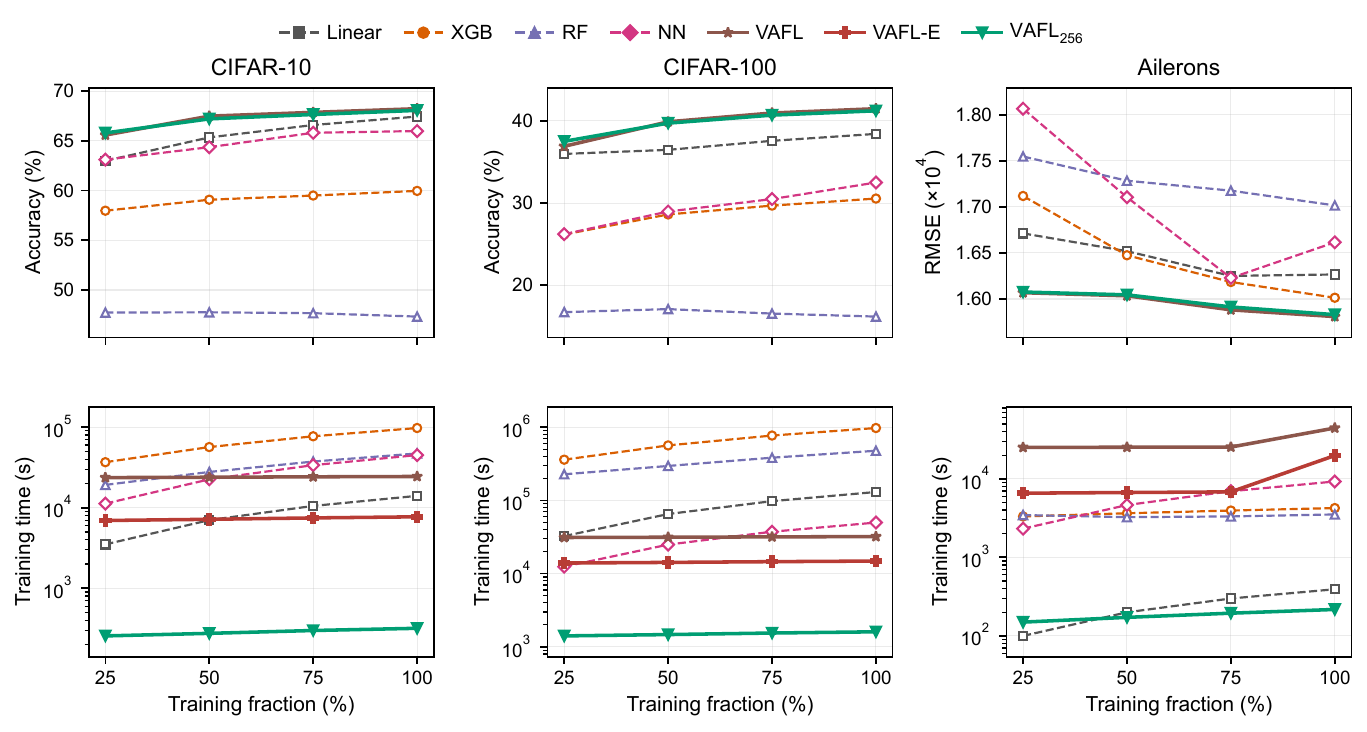}
\caption{Prediction and training time versus training fraction. Reduced VAFL uses LC on CIFAR and SP on Ailerons.}
\label{fig:exp-training-size}
\Description{Six panels compare four baselines with VAFL variants on three datasets. Top panels report mean test performance, and bottom panels report training time on logarithmic axes. Legends are outside the axes and no error bars are plotted.}
\end{figure}

Figure~\ref{fig:exp-training-size} shows the effect of increasing the labeled training set. From 25\% to 100\%, original VAFL accuracy increases from 65.58\% to 68.23\% on CIFAR-10 and from 36.91\% to 41.49\% on CIFAR-100. The corresponding LC$_{256}$ endpoints are 65.79\%/68.06\% and 37.51\%/41.23\%. LC can slightly exceed the fixed-penalty full-dimensional reference with fewer training records. This comparison also includes its validation-based regularization selection. At full data it remains close to the original predictor.

On Ailerons, original VAFL RMSE changes from 1.607 to 1.581, while SP$_{256}$ changes from 1.608 to 1.583. The two curves remain close without requiring the projected head to reproduce the full-dimensional coefficients. The four baseline curves provide the same-subset comparison. Individual intermediate points need not improve monotonically.

The lower panels show a different scaling pattern for the analytic and iterative heads. For the fourfold increase in training records, original VAFL training-time ratios are 1.03$\times$, 1.03$\times$, 1.77$\times$ on CIFAR-10, CIFAR-100 and Ailerons, respectively. The corresponding Linear ratios are 4.00$\times$, 4.00$\times$, 3.93$\times$. Additional records change feature and target products, while the feature-space solve retains its matrix dimensions. VAFL-E reduces full-dimensional solve work, and LC/SP reduce the collaborative width further. These gains do not imply that the reduced head is the fastest method at every training fraction. The Ailerons step at full data coincides with the recorded conditioning threshold activating the direct-Gram fallback. The feature width itself is unchanged.

\begin{figure}[H]
\centering
\includegraphics[width=\linewidth]{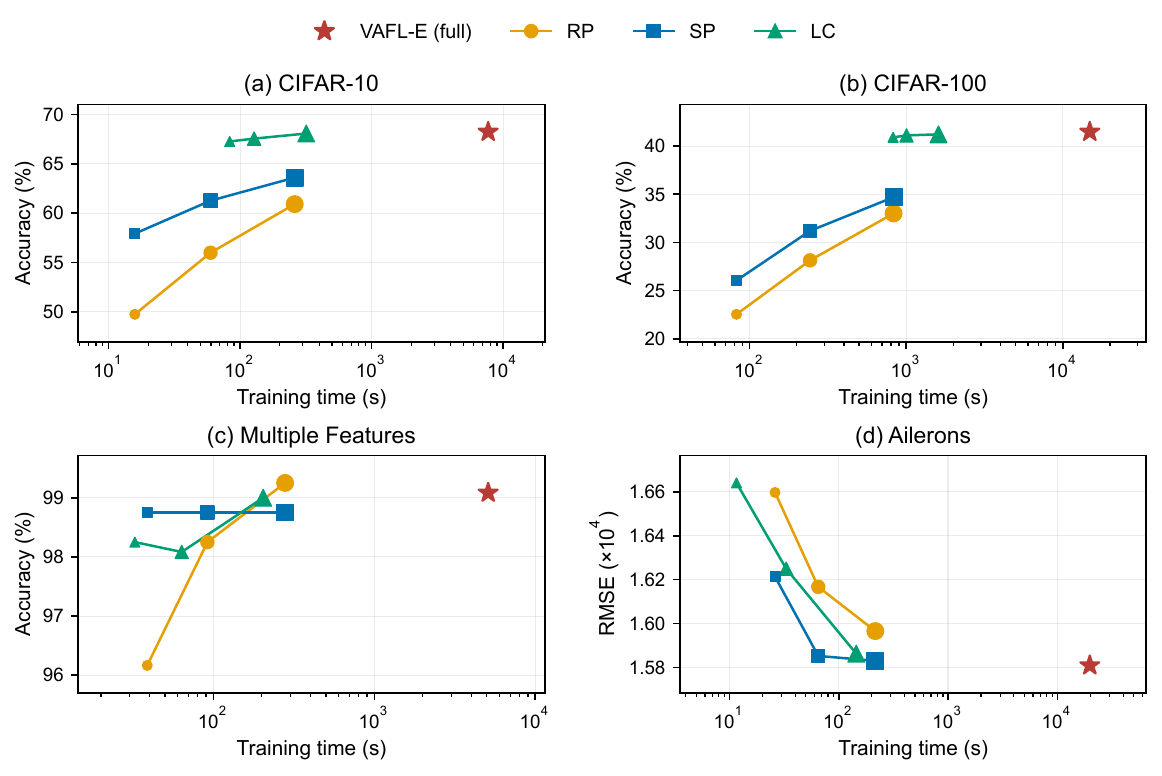}
\caption{Prediction versus training time. Along each RP, SP and LC curve, the three points correspond to budgets $m=64,128,256$ from left to right.}
\label{fig:acceleration-tradeoff}
\Description{Four panels compare full-dimensional VAFL-E with RP, SP and LC. Along each reduced-strategy curve, the three points from left to right use total budgets 64, 128 and 256. The upper legend identifies methods. All observed points are retained, and time axes are logarithmic.}
\end{figure}

\subsection{RQ3: Evaluation of Acceleration Strategies}
\label{sec:exp-acceleration}
\label{sec:exp-efficiency}
Figure~\ref{fig:acceleration-tradeoff} compares VAFL-E with RP, SP and LC on CIFAR-10/100, MF and Ailerons at $K=2$. The reduced strategies use matched total non-bias budgets of 64, 128 and 256. RP and SP use the same equivalent block-factor solve. LC additionally retains full-dimensional local target products and local coefficient application. Its workload is therefore not that of a purely projected head.

The strategies exhibit different prediction--cost trade-offs. At $m=256$, RP, SP and LC attain 60.92\%, 63.60\% and 68.06\%, respectively, on CIFAR-10. Their CIFAR-100 accuracies are 33.01\%, 34.70\% and 41.23\%. LC preserves full local components that pure projection discards. Relative to VAFL-E, the 256-dimensional training-time reductions are 95.9\% and 89.2\%, with prediction losses below 0.3 percentage points.

The tabular results motivate retaining SP as the default reduced variant rather than applying LC everywhere. On MF, RP$_{256}$ reaches 99.25\%, whereas SP gives 98.75\% at all three widths. No projection dominates every operating point. LC can deteriorate substantially with more parties, especially on Ailerons at small budgets. Thus the two-party image result does not imply universal accuracy preservation. The full dataset--party--budget comparisons and their analysis remain in Appendix~H.

\subsection{RQ4: Scalability with the Number of Parties}
\label{sec:exp-scaling}
We vary $K\in\{2,4,8,16\}$ on the three tabular datasets while keeping the global representation and records fixed. Diabetes omits $K=16$ because it has only ten input features. Original VAFL maintains 99.08\% on MF and RMSEs of 56.442 and 1.581 on Diabetes and Ailerons. These are full-dimensional partition-invariance controls. Low-rank predictions can change with the partition. For CIFAR, changing $K$ also changes the spatial crops and hence the supplied features, so its rows are not a pure repartitioning experiment.

\begin{figure}[H]
\centering
\includegraphics[width=\linewidth]{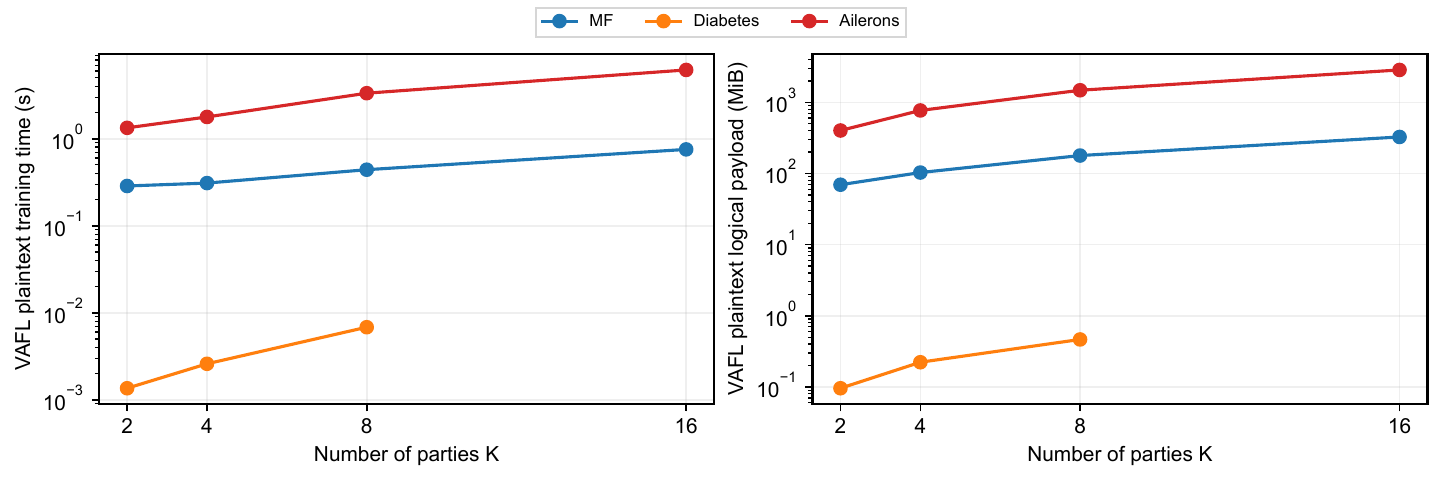}
\caption{Plaintext-protocol training time and logical payload versus party count. Diabetes omits $K=16$ because it has only ten features.}
\label{fig:exp-scaling}
\Description{Two logarithmic panels show original plaintext-protocol training time and logical payload on three tabular datasets.}
\end{figure}

The retained plaintext-protocol control in Figure~\ref{fig:exp-scaling} rises from 0.286 to 0.753 seconds on MF and from 1.33 to 6.16 seconds on Ailerons as $K$ grows from 2 to 16. Additional cross-party interactions increase logical payload. These plaintext quantities use the original implementation.

\subsection{RQ5: Sensitivity Analysis}
\label{sec:exp-robustness}
We examine regularization and labeled training samples at $K=2$ on all five datasets. The full-dimensional VAFL curve also represents VAFL-E. Reduced variants use LC on CIFAR and SP on the tabular datasets at $m=64,128,256$. Diabetes retains its native features and is shown once as an identity control. Figures report mean predictive performance without error bars.

\begin{figure}[H]
\centering
\includegraphics[width=\linewidth]{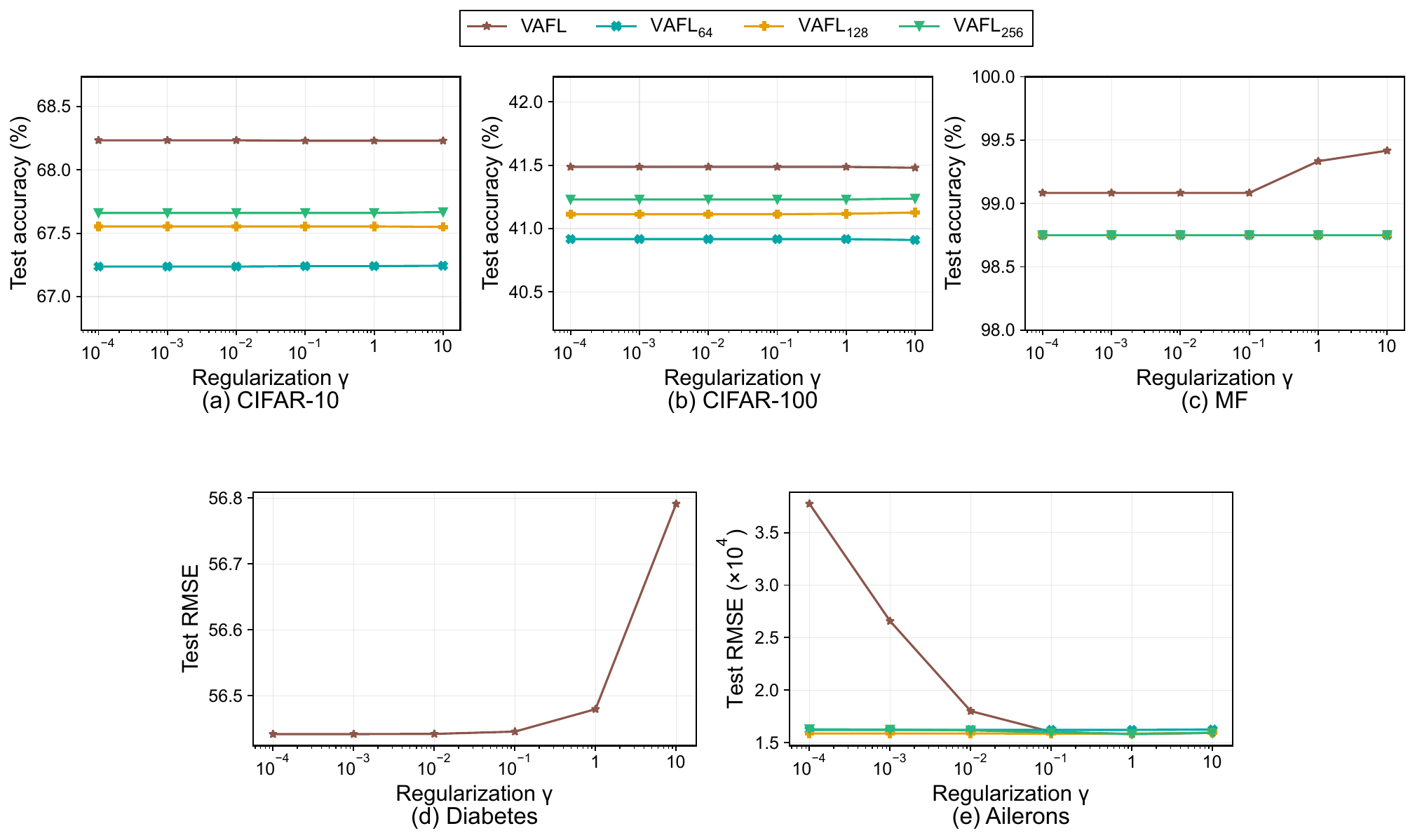}
\caption{Sensitivity to regularization at $K=2$. Reduced VAFL uses LC on CIFAR and SP on tabular data. Coincident curves overlap.}
\label{fig:exp-regularization}
\label{fig:rq4-compressed-regularization}
\Description{Five panels compare mean predictive performance across six penalties. CIFAR reduced variants use LC, while MF and Ailerons use SP. Diabetes is an identity control.}
\end{figure}

\paragraph{\textbf{\upshape Regularization.}} We sweep $\gamma\in\{10^{-4},10^{-3},10^{-2},10^{-1},1,10\}$ with the training data fixed. Each point applies the same penalty to the full and reduced systems. This six-value diagnostic grid is narrower than the validation grid used for CIFAR LC in the main table. Figure~\ref{fig:exp-regularization} shows that the CIFAR curves vary by less than 0.02 percentage points over this range. On MF, full-dimensional accuracy changes from 99.08\% to 99.42\%, while the three SP variants share 98.75\% and overlap in the plot. On Ailerons, full-dimensional RMSE decreases from 3.775 to 1.581 as the penalty increases to one. SP$_{256}$ changes from 1.627 to 1.583. Regularization therefore remains relevant to regression and does not have the same effect across representations.

\begin{figure}[H]
\centering
\includegraphics[width=\linewidth]{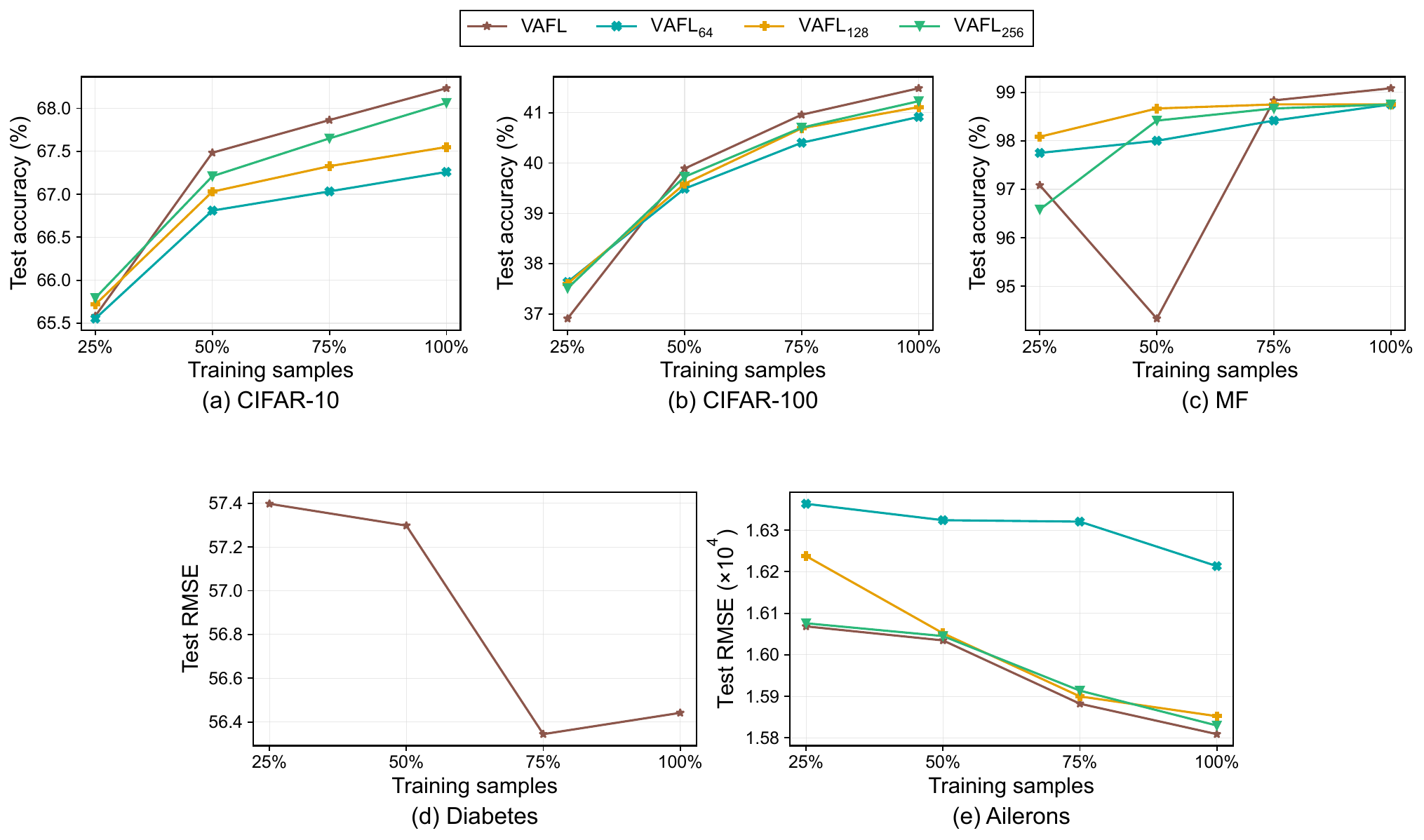}
\caption{Sensitivity to training samples.}
\label{fig:exp-training-sensitivity}
\label{fig:rq4-compressed-training-fraction}
\Description{Five panels compare mean predictive performance at four nested training fractions, with LC image variants and SP tabular variants. All measured points, including non-monotonic results, are retained.}
\end{figure}

\paragraph{\textbf{\upshape Training samples.}} We vary the labeled head-training fraction over 25\%, 50\%, 75\% and 100\%, retaining fixed validation and test partitions. The original head and SP keep their dataset-specific penalties. CIFAR LC selects a common penalty across splits by validation for each fraction and budget. Figure~\ref{fig:exp-training-sensitivity} extends the comparison in Figure~\ref{fig:exp-training-size} to all five datasets and all three budgets. CIFAR-10 LC$_{64}$ increases from 65.55\% to 67.26\%, while CIFAR-100 LC$_{64}$ increases from 37.63\% to 40.92\%. At full data, LC$_{256}$ stays closest to the original image head among the displayed reduced budgets.

The tabular curves also retain non-monotonic outcomes. With the fixed original MF penalty, full-dimensional accuracy at 25\%, 50\%, 75\% and 100\% is 97.08\%, 94.33\%, 98.83\%, 99.08\%. More head-training records do not ensure a monotonic improvement at a fixed penalty. On Ailerons, SP$_{256}$ RMSE changes from 1.608 to 1.583, close to the full-dimensional endpoints 1.607 and 1.581. These results describe sensitivity under the specified representations and parameter rules, without asserting accuracy preservation for every budget.

\subsection{RQ6: Compatibility with Frozen Backbones}
\label{sec:exp-versatility}
We compare VAFL with Linear, XGB, RF and NN using ResNet18, MobileNetV2, ShuffleNetV2 ($1.0\times$) and ViT-MAE-base on CIFAR-10 and CIFAR-100 at $K=2$. All methods use the same 45,000 training, 5,000 validation and 10,000 test images within each dataset, and share the uncompressed local features for each backbone. We retain $\gamma=10^{-4}$ for VAFL, 30 Adam epochs with learning rate 0.005 and batch size 256 for Linear and NN, and 50 trees with maximum depth 6 for XGB and RF. Further baseline settings are given in Appendix~F.2. In the listed order, the frozen encoders contain 11.18, 2.22, 1.25 and 85.80 million parameters, with corresponding float32 parameter storage requirements of 42.64, 8.48, 4.78 and 327.30~MiB.

\begin{table}[htb]
\centering
\caption{Frozen-backbone accuracy (\%) on CIFAR-10 and CIFAR-100, reported as mean $\pm$ SEM. Bold marks the best mean.}
\label{tab:exp-backbones}
\begingroup\normalfont\normalsize
\setlength{\heavyrulewidth}{0.6pt}
\setlength{\lightrulewidth}{0.4pt}
\setlength{\tabcolsep}{5pt}\renewcommand{\arraystretch}{1.10}
\begin{tabular}{c|lccccc}
\toprule
Dataset & Frozen backbone & Linear & XGB & RF & NN & VAFL \\
\midrule
\multirow[c]{4}{*}{CIFAR-10} & ResNet18 & $67.44_{\pm 0.44}$ & $59.97_{\pm 0.11}$ & $47.33_{\pm 0.02}$ & $65.99_{\pm 0.13}$ & $\mathbf{68.23_{\pm 0.01}}$ \\
 & MobileNetV2 & $66.87_{\pm 0.12}$ & $61.83_{\pm 0.17}$ & $50.11_{\pm 0.06}$ & $69.08_{\pm 0.44}$ & $\mathbf{71.42_{\pm 0.05}}$ \\
 & ShuffleNetV2 ($1.0\times$) & $67.04_{\pm 0.10}$ & $60.89_{\pm 0.21}$ & $48.24_{\pm 0.03}$ & $68.31_{\pm 0.06}$ & $\mathbf{71.03_{\pm 0.07}}$ \\
 & ViT-MAE-B & $81.32_{\pm 0.37}$ & $68.85_{\pm 0.14}$ & $52.17_{\pm 0.08}$ & $81.96_{\pm 0.06}$ & $\mathbf{82.64_{\pm 0.03}}$ \\
\midrule
\multirow[c]{4}{*}{CIFAR-100} & ResNet18 & $38.42_{\pm 0.07}$ & $30.55_{\pm 0.15}$ & $16.18_{\pm 0.24}$ & $32.50_{\pm 0.12}$ & $\mathbf{41.49_{\pm 0.08}}$ \\
 & MobileNetV2 & $40.94_{\pm 0.11}$ & $32.23_{\pm 0.18}$ & $16.70_{\pm 0.06}$ & $33.31_{\pm 0.19}$ & $\mathbf{45.25_{\pm 0.06}}$ \\
 & ShuffleNetV2 ($1.0\times$) & $39.24_{\pm 0.26}$ & $31.33_{\pm 0.13}$ & $14.00_{\pm 0.25}$ & $31.74_{\pm 0.29}$ & $\mathbf{42.91_{\pm 0.08}}$ \\
 & ViT-MAE-B & $57.43_{\pm 0.29}$ & $38.61_{\pm 0.04}$ & $17.98_{\pm 0.21}$ & $50.12_{\pm 0.44}$ & $\mathbf{58.23_{\pm 0.08}}$ \\
\bottomrule
\end{tabular}
\endgroup
\end{table}

As shown in Table~\ref{tab:exp-backbones}, VAFL achieves the highest observed mean accuracy in all eight dataset--backbone configurations. On CIFAR-10, its margins over the strongest baselines range from 0.68 to 2.71 percentage points. On CIFAR-100, its margins over the strongest baselines range from 0.81 to 4.31 percentage points. With ViT-MAE-B, VAFL reaches 82.64\% and 58.23\% on CIFAR-10 and CIFAR-100, respectively. MobileNetV2 and ShuffleNetV2 give higher VAFL accuracy than ResNet18 on both datasets while reducing local encoder storage. These results demonstrate VAFL's compatibility with both lightweight CNN and transformer representations while retaining the same analytic training procedure across all four backbones.

\FloatBarrier

\section{Conclusion and Discussion for Future Work}
\label{Conclusion}

VAFL fits classification and regression heads on vertically partitioned fixed representations through analytic matrix computation. Building on the shared foundations of Ridge fitting and feature-block recursion, we distinguish three execution choices: full-dimensional equivalent solving, local spectral projection with lifting, and full-dimensional local fitting with low-rank cross-party correction. The equivalent solve preserves the original optimum in exact arithmetic. The two reduced strategies address width-dependent overhead, with LC retaining useful local components that pure projection omits.

Measured plaintext results on five benchmarks support the predictive usefulness of the analytic head. On the two CIFAR datasets, LC substantially improves over matched-width projection, especially with two parties, but many-party compression still loses accuracy. The dedicated scaling and acceleration experiments characterize fitting cost and the computational benefits of the proposed strategies. Approximation losses remain significant in some configurations. Full-dimensional fitting remains the reference when retaining the original predictor is required.
Future work includes conditioning-aware collaboration subspaces and fair regularization-budget comparisons with direct vertical Ridge solvers. Deeper analytic representations~\cite{DeepAFL} may improve the linear head's modeling capacity. Extending the workflow to hybrid sample/feature partitioning is promising~\cite{NC-fed-joint,Mobi-fed-joint}.

\bibliographystyle{ACM-Reference-Format}
\bibliography{reference}

\end{document}